\documentclass[conference]{IEEEtran}
\IEEEoverridecommandlockouts
\usepackage{cite}
\usepackage{amsmath,amssymb,amsfonts}
\usepackage{algorithmic}
\usepackage{graphicx}
\usepackage{textcomp}
\usepackage{xcolor}
\usepackage{multirow}
\usepackage{subfigure}
\usepackage{booktabs}
\usepackage{algorithm}
\usepackage{algorithmic}
\usepackage{hyperref}
\usepackage{comment}

\def\BibTeX{{\rm B\kern-.05em{\sc i\kern-.025em b}\kern-.08em
    T\kern-.1667em\lower.7ex\hbox{E}\kern-.125emX}}
\begin{document}

\title{KGLink: A column type annotation method that combines knowledge graph and pre-trained language model}

\author{\IEEEauthorblockN{1\textsuperscript{st} Yubo Wang}
\IEEEauthorblockA{
\textit{HKUST}\\
Hong Kong, China \\
ywangnx@cse.ust.hk}
\and
\IEEEauthorblockN{2\textsuperscript{nd} Hao Xin}
\IEEEauthorblockA{
\textit{HKUST}\\
Hong Kong, China \\
hxinaa@cse.ust.hk}
\and
\IEEEauthorblockN{3\textsuperscript{rd} Lei Chen}
\IEEEauthorblockA{
\textit{HKUST(GZ) \& HKUST}\\
Guangzhou, China \\
leichen@cse.ust.hk}
}

\maketitle

\begin{abstract}
The semantic annotation of tabular data plays a crucial role in various downstream tasks. Previous research has proposed knowledge graph (KG)-based and deep learning-based methods, each with its inherent limitations. KG-based methods encounter difficulties annotating columns when there is no match for column cells in the KG. Moreover, KG-based methods can provide multiple predictions for one column, making it challenging to determine the semantic type with the most suitable granularity for the dataset. This type granularity issue limits their scalability.

On the other hand, deep learning-based methods face challenges related to the valuable context missing issue. This occurs when the information within the table is insufficient for determining the correct column type.

This paper presents KGLink, a method that combines WikiData KG information with a pre-trained deep learning language model for table column annotation, effectively addressing both type granularity and valuable context missing issues. Through comprehensive experiments on widely used tabular datasets encompassing numeric and string columns with varying type granularity, we showcase the effectiveness and efficiency of KGLink. By leveraging the strengths of KGLink, we successfully surmount challenges related to type granularity and valuable context issues, establishing it as a robust solution for the semantic annotation of tabular data.
\end{abstract}

\begin{IEEEkeywords}
knowledge graph, column type annotation, data mining.
\end{IEEEkeywords}


\section{Introduction}
Table column annotation is a crucial task, and many companies develop commercial systems like Google Data Studio \cite{GDS}, Microsoft Power BI \cite{MPBI}, and Tableau \cite{tableau}, to understand tables and perform downstream tasks such as data visualization. Current column-type annotation methods can be categorized into three main groups: knowledge graph (KG)-based methods, deep learning-based methods, and hybrid methods. 

\begin{figure*}[t]
 \centering
 \subfigure[\small{Table with type granularity issue}]{
  \label{fig:tg_issue}
  \includegraphics[scale=0.65]{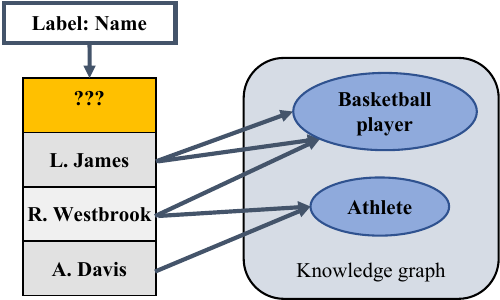}
 }\hspace{10mm}
  \subfigure[\small{Table with valuable context missing issue}]{
  \label{fig:DL_NotWell}
  \includegraphics[scale=0.65]{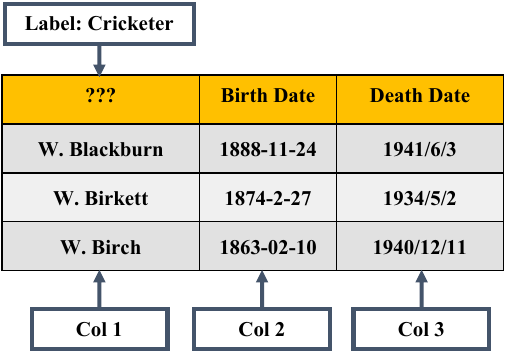}
 }
 
 \caption{Figure (a) and (b) present examples for the type granularity issue and valuable context missing issue, respectively. In Figure (a), for a column of basketball player names, the column type fetch from the KG could be \textit{Athlete}, or \textit{Basketball player}. These two types could all be correct types. However, in this dataset, the ground truth type (label) desired for this column could be \textit{Name} since no finer granularity types such as \textit{Athlete} exist in the dataset labels. A granularity gap exists between type \textit{Athlete} or \textit{Basketball player} and \textit{Name}. Figure (b) highlights valuable context missing issues, posing challenges for deep learning-based models in performing column annotation tasks. The context information from columns two and three, which are irrelevant to the column type annotation task on column one, fails to provide any information close to the ground truth label: \textit{Cricketer}. Consequently, it becomes arduous for deep learning-based models to accurately annotate the target column.}
 \vspace{-0.2cm}
 \label{fig:m_w_F1}
\end{figure*}

The most recent KG-based method is MTab \cite{2021_mtab4wikidata}, which leverages information from KGs such as WikiData \cite{vrandevcic2014wikidata} and DBpedia \cite{lehmann2015dbpedia}. These methods typically rely on rules or statistic-based techniques like TF-IDF (Term Frequency-Inverse Document Frequency) and heavily depend on the relatedness between the table and the KG. Firstly, in real-world scenarios, there are usually numerous suitable types in KG with different type granularity for a certain column, but not all of those types are necessarily the ones we desire. As illustrated in Figure \autoref{fig:tg_issue}, KG-based methods might generate candidate types such as \textit{Athlete} or \textit{Basketball player}, which possess closely related and semantically correct meanings with a finer granularity than the ground truth type \textit{Name}. Despite their relevance, these candidate types do not directly contribute to the prediction because they do not match the expected \textit{Name}, which does not even exist in the type hierarchy of types \textit{Athlete} or \textit{Basketball player}. We refer to this misalignment in type granularity as the \lq type granularity gap'. This gap does not only impacts model performance but also constrains the scalability of purely KG-based methods, necessitating a way to enhance the contribution of correct candidate types with similar semantic meanings to the ground truth type. Secondly, KG-based methods face challenges in making predictions when they cannot find relevant information for columns in the KG, particularly for numeric columns, since they are unsuitable to be linked to the KG.

To enhance scalability, various deep learning-based methods have been proposed, including TaBERT \cite{yin2020tabert} and Doduo \cite{suhara2022annotating}. These methods generate representation vectors for columns during training, showcasing improved scalability. Nevertheless, they grapple with valuable context missing issues. This implies that the input table often lacks sufficient semantic information related to the predicted column type of the dataset. In Figure \autoref{fig:DL_NotWell}, for example, the target column (predicted column) lacks essential context information, such as \textit{Sports Teams} or \textit{Position in Team}, which is crucial for accurately identifying the column type as \textit{Cricketer}. Notably, neither \textit{Birth Date} nor \textit{Death Date} is helpful for the deep learning model prediction. Conversely, although pre-trained language model (PLM)-based methods, such as Doduo \cite{suhara2022annotating}, achieve decent performance in the column annotation task, the PLMs’ input sequence length limitations could impede their scalability, especially on tables with numerous rows or columns.

\begin{figure}[t]
  \centering
  \includegraphics[width=\linewidth]{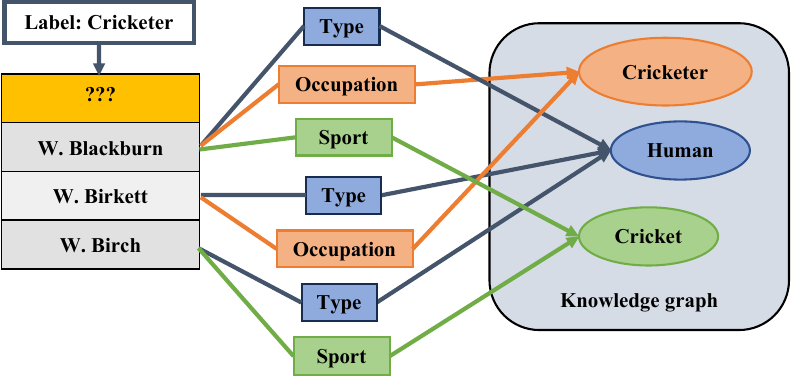}
  \caption{An example from the SemTab dataset, if we only consider the \textit{type} attribute, we would only obtain \textit{Human} as the candidate type from the KG. This approach would overlook \textit{Cricketer} and \textit{Cricket}, which offer a finer granularity than \textit{Human} and also provide valuable information for the column type annotation task.}
  \label{fig:1_hop}
 \vspace{-0.4cm}
\end{figure}

Given the drawbacks of solely KG-based models and solely deep learning-based models, researchers have explored hybrid models. Some existing hybrids, like ColNet \cite{chen2019colnet} and its more generalized version HNN \cite{chen2019learning}, aim to integrate KG information with deep learning models. However, as illustrated in \autoref{fig:1_hop}, these models employ overly simplistic strategies in selecting information from the KG and often fall short of fully leveraging table structure and KG information. Firstly, HNN generates column embeddings solely based on the entity type from a single cell in that column, disregarding interactions among columns and rows and neglecting crucial table structure information. Secondly, when selecting types for entities, HNN only considers types that belong to the KG-provided \textit{type} attribute of entities. This approach would sometimes exclude the desired types, as those types may not appear in the \textit{type} attributes. The heavy reliance on the quality of a single cell's linkage could easily introduce noise, and depending solely on KG-provided types could cause the model to overlook many other valuable pieces of information in the KG. These factors could adversely impact the model's performance and generalization abilities. (see \autoref{KB}).

The examples provided above indicate that effectively integrating complex KG information into deep learning-based models is a future research direction that still presents persistent challenges and is worth exploring.

Firstly, calling back numerous KG entities for one cell in the table may not always be helpful, as some incorrect or missing entity linkages in the KG can negatively affect the model prediction. 

Secondly, even if the true entity of one cell is retrieved, identifying the type that best aligns with the semantic meaning of a dataset is challenging, as these types may not be readily associated with the type attribute of the retrieved entities. 

Lastly, the scalability of PLM-based models remains questionable on large tables due to input sequence length limits. The incorporation of additional KG information in the input sequence of PLMs could exacerbate this problem, requiring an approach to reduce the size of the tables fed to PLMs.


To overcome these challenges, we introduce KGLink, an approach that integrates KG information with deep learning models for column type annotation. In the KGLink process, tables are serialized, and the PLM utilizes both KG information and the serialized table to generate encodings for the table and feature vectors. This enhances the overall prediction process.

Firstly, the KGLink process involves searching for mentions of table cells in the external KG, then retrieving and filtering entities based on the table structure to retain the best-matched entities. Subsequently, to avoid a situation where the filter excludes all information from KG, we generate a feature vector for each column using KG information. These vectors serve as supplementary context information, addressing the valuable context missing issue illustrated in \autoref{fig:DL_NotWell}.

Secondly, to identify the best type align with the entity, we propose a candidate type representation generation task. It aims to leverage the generalization ability of PLM and thereby resolving the type granularity issue depicted in \autoref{fig:tg_issue}.

Lastly, we introduce the row linking score for filtering rows based on their linkage quality. This measure aims to reduce the size of the tables fed to PLMs while preserving prediction effectiveness.

Furthermore, even for columns whose cells have no linkage with KG, such as numeric columns, our model can still provide reasonable predictions due to the prior knowledge from the PLM.

\noindent Our contributions are summarized as follows:
\begin{itemize}

\item We proposed KGLink, a method for the column-type prediction task that effectively combines KG information with prior knowledge from a pre-trained language model (PLM). This approach addresses the type granularity issue in KG-based methods and the valuable context missing issue in deep learning methods.

\item We introduced an overlapping score to filter out KG-retrieved entities based on the table structure. Additionally, we utilized a feature vector generated from KG-retrieved information to ensure that some information remains after the filtering process.

\item We introduced the candidate-type representation generation task as a sub-task to further address the type granularity issue and improve model performance (Please refer to \autoref{tdl} for details).



\item We have made the code of our model available on GitHub\footnote{https://github.com/Wyb0627/KBLink}, along with the modified VizNet \cite{zhang2020sato} and SemTab \cite{hassanzadeh_oktie_2019_3518539} datasets, where entities have been linked to the WikiData \cite{vrandevcic2014wikidata} KG.
\end{itemize}

\section{Related Work}
As discussed in the previous section, column-type annotation methods can be broadly categorized into two main types:

\noindent \textbf{Knowledge graph based methods:} These methods, such as MTab \cite{2021_mtab4wikidata}, ADOG \cite{oliveira2019adog}, and JenTab \cite{abdelmageed2020jentab}, aim to establish connections between table cells and external knowledge graphs (KGs). They utilize rule-based or statistic-based filters to enhance linking accuracy. While some methods perform well on KG-extracted datasets with exact entity matches, others use the KG as an external information source without requiring exact matches. For example, Khurana et al. proposed $C^2$ \cite{khurana2021semantic}, which integrates multiple KGs such as Wikidata and DBpedia to form the data lake. It employs maximum likelihood estimation to predict column types by leveraging information from the external KG.

\noindent \textbf{Deep learning-based methods:} These methods approach column type annotation as multi-class classification task. Hulsebos et al. \cite{hulsebos2019sherlock} introduced Sherlock, a deep learning based model that utilized word embeddings and global statistics features to predict column types. Building upon Sherlock, Zhang et al. proposed Sato \cite{zhang2020sato}, which incorporated the table context by integrating an LDA model to generate a global vector representing the table's context and capturing the correlation between its columns and cells. Another deep learning-based model is Tabbie \cite{iida2021tabbie}, introduced by Iida et al., which is a PLM that takes the table structure into consideration. It leverages stacked row and column transformers to generate table cell embeddings and perform the column type annotation task. Wang et al. proposed TCN \cite{wang2021tcn}, a convolutional network-based PLM that utilizes information within the table and among similar domain tables to predict column types and relations. Recently, various PLMs such as BERT \cite{devlin2018bert} and RoBERTa \cite{liu2019roberta} have gained popularity in table understanding tasks as useful tools. Yin et al. introduced TaBERT \cite{yin2020tabert}, which builds upon the BERT PLM and achieves promising performance in table understanding tasks. Doduo \cite{suhara2022annotating}, proposed by Suhara et al., is another deep learning-based model that utilizes a stacked transformer structure and BERT. It integrates the column type annotation task with the column relation prediction task, allowing them to benefit from each other. TURL \cite{deng2022turl}, introduced by Deng et al., employs a visibility matrix to mask out structurally irrelevant table components and enhance the BERT prediction. Sudowoodo \cite{wang2022sudowoodo}, proposed by Wang et al., applies k-means clustering and contrastive learning techniques with the RoBERTa PLM for unsupervised or semi-supervised column type annotation, greatly lower down the training data amount requirement. Sun et al. developed RECA \cite{sun2023reca}, which leverages inter-table information within the training dataset and BERT to achieve state-of-the-art column type annotation performance.
\\ 

\begin{figure*}[ht]
  \centering
  \includegraphics[width=0.9\linewidth]{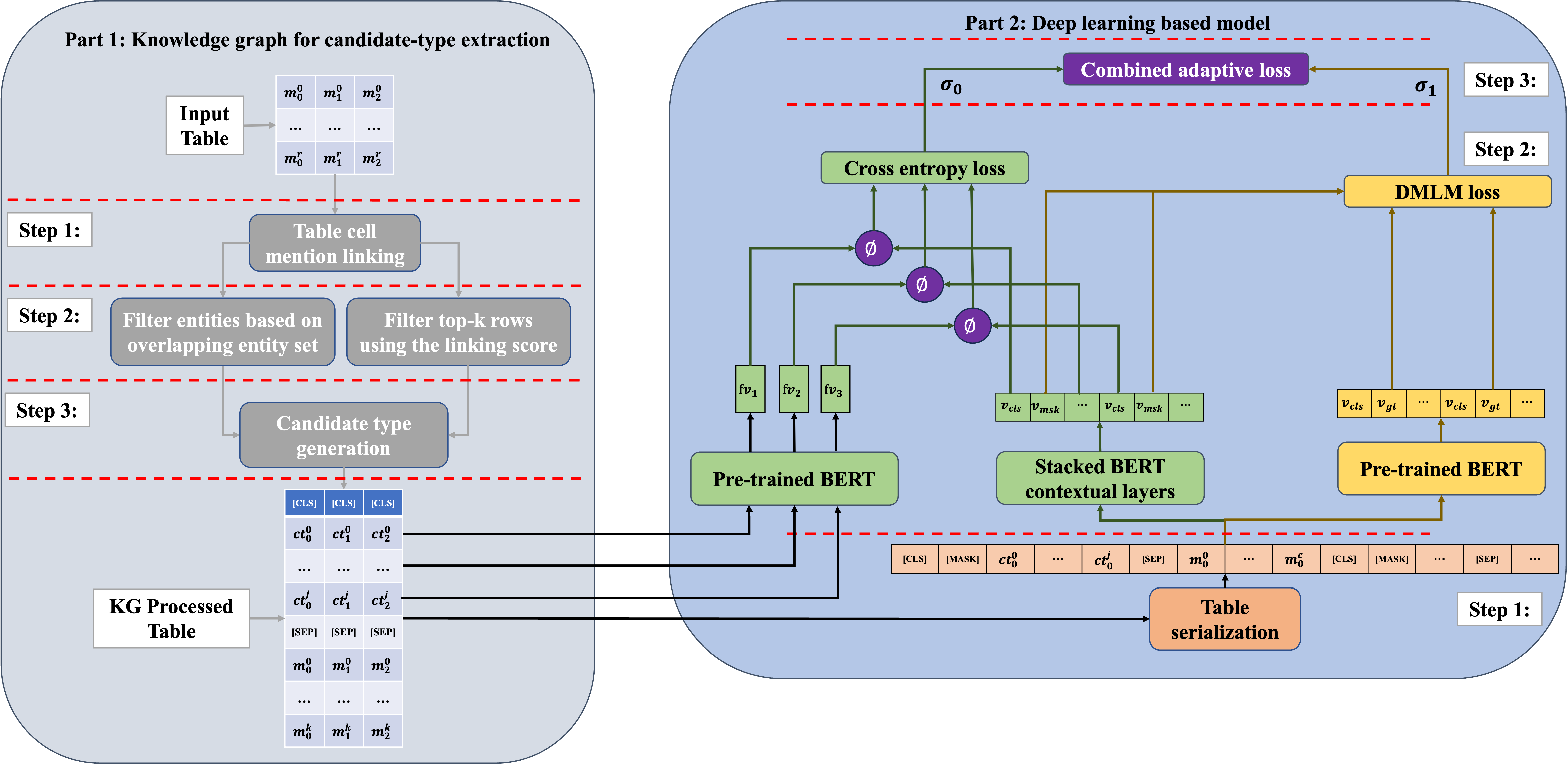}
  \caption{Overview of KGLink's model structure: KGLink integrates a knowledge graph (KG) component designed to filter out inappropriate numeric or date table cells. It proceeds with identifying $j$ candidate types $ct_0, \dots ct_j$ for each column. Following this, the table rows undergo sorting based on their linkage quality. The processed table incorporates labels for candidate-type entities in each column. This table is then serialized by the deep learning component in part 2, where feature vectors for each column are generated using information fetched from the KG in part 1. KGLink introduces the column type representation generation task as an subtask to further enhance prediction performance. This task generates a representation vector for each predicted column based on its cell and KG-extracted information. It then aims to optimize the gap between this vector and the representation vector of the column's label in the dataset. To optimize accuracy, the model utilizes a combined adaptive loss with trainable weights $\sigma_0$ and $\sigma_1$. Collectively, these elements contribute to the improved accuracy of column type predictions.}
  \label{fig:KGLink}
  \vspace{-0.3cm}
\end{figure*}

\noindent \textbf{Hybrid methods:} Several earlier studies have integrated knowledge graphs (KGs) with machine learning models. Chen et al. introduced ColNet \cite{chen2019colnet}, a CNN-based approach that utilizes the DBpedia KG to generate synthetic columns, thereby augmenting training samples. The method leverages the KG to generate candidate types, aiding in CNN predictions. ColNet employs a custom binary CNN classifier for each candidate class, predicting whether the cells in a column belong to a specific class. However, the approach of training separate classifiers for each type in ColNet is time-consuming and limits scalability.

Expanding on ColNet, Chen et al. introduced HNN \cite{chen2019learning}, an enhanced version that incorporates inter-column semantic information and utilizes the main cell entity's property in the KG for prediction assistance. However, HNN has limitations as it only considers linking the first cell of each target column to the KG and exclusively focuses on types within the KG-provided \textit{type} attribute of entities. This simplistic approach could introduce noise if the linkages of the sole cell are incorrect. Additionally, retrieving only the \textit{type} attribute of entities overlooks many other valuable pieces of information present in the KG.

In this paper, we empirically compare KGLink with five baseline methods including TaBERT \cite{yin2020tabert}, Doduo \cite{suhara2022annotating}, HNN \cite{chen2019learning}, Sudowoodo \cite{wang2022sudowoodo}, and RECA \cite{sun2023reca}. Our experiments are based on the modified VizNet\footnote{https://github.com/megagonlabs/sato/tree/master/table\_data} \cite{zhang2020sato} and SemTab\footnote{https://zenodo.org/record/3518531} \cite{hassanzadeh_oktie_2019_3518539} datasets.

\section{Methdology} \label{model}
The type granularity and valuable context missing issues can significantly affect the prediction accuracy of column type annotation models. In response to these challenges, we introduce a hybrid approach that melds KG information with a pre-trained language model (PLM) based on deep learning.

Our model, illustrated in \autoref{fig:KGLink}, is divided into two main parts: Part 1 is centered around extracting candidate types from KG, while Part 2 employs a deep learning model to perform the tasks of column type annotation and column representation generation. In Part 1, the initial table taken from the dataset, featuring the target column, is used as input, resulting in a processed table with KG-derived candidate types. In Part 2, this processed table serves as input for generating predictions for the column type annotation and column representation generation tasks. Further detailed explanations of these components will be presented in the subsequent sections.

\subsection{Part 1: Knowledge graph for candidate type extraction}\label{KB} 

\begin{figure}[t]
  \centering
\includegraphics[width=\linewidth]{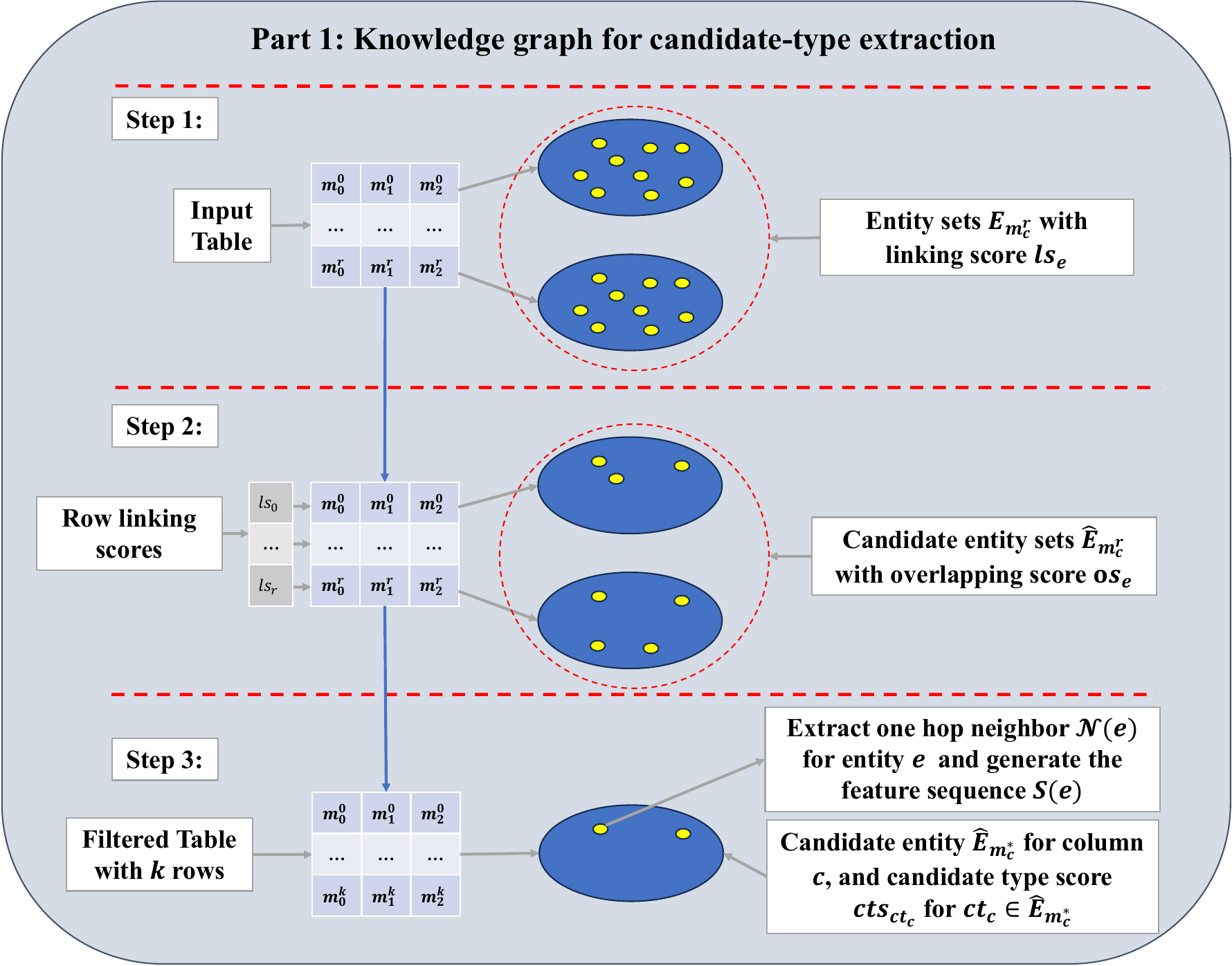}
  \caption{Overview of the KG candidate type extraction process: We break down this procedure into three steps. This division is designed to minimize noise from the KG, generate the feature sequence, and optimize the table for enhanced predictions in the subsequent deep learning-based model.}
  \vspace{-0.3cm}
  \label{fig:KG}
\end{figure}

In this section, we delineate our approach to tackle the valuable context missing issue by leveraging KG information. To address this challenge in a table $T$ with $n$ columns and $l$ rows, we adopt a three-step process, as depicted in \autoref{fig:KG}.




By adhering to this comprehensive approach, we effectively leverage KG information to mitigate valuable context missing, thereby strengthening the predictive capabilities of the deep learning model.

\noindent \textbf{Step 1: Table cell mention linking:} 
In this step, our objective is to obtain a KG entity set $E_{m_c^r}$ for each cell mention $m_c^r$.
To establish a connection between a cell mention $m$ within the table and the KG, we link $m$ to the KG and retrieve its associated entity set $E_m$. This connection is referred to as a \textit{link}, and the degree of correlation between the cell mention and the entity label is quantified as the \textit{linking score} ($ls$) of this \textit{link}. Within our approach, we employ the BM25 score \cite{robertson2009probabilistic} (BM, denoting Best Matching) as the linking score for table cell mentions to their corresponding entities within the KG:

\vspace{-0.5cm}
\begin{equation}\label{bm25}
ls_e=\sum_{l=1}^n{IDF(w_l)} \cdot \frac{f(w_l,E) \cdot (k_1+1)}{f(w_l,e)+k_1\cdot\Big(1-b+b\cdot \frac{|e|}{avgwl}\Big)},
\end{equation}
where $e$ signifies the entities within the KG entity set $E$. Given that a table cell mention could encompass multiple words, $m=\{w_1,w_2,...,w_l\}$ refers to the search query containing the words extracted from the table cell's mention. $f(w_l, E)$ indicates the count of occurrences of the word $w_l$ in the entity $E$, $|E|$ denotes the length of the entity $E$ in terms of words, and $avgwl$ represents the average entity length in words across the complete indexed KG. The parameters $k_1$ and $b$ are adjustable parameters. The notation $IDF(w_l)$ stands for the inverse document frequency weight associated with the word $w_l$:

\vspace{-0.1cm}
\begin{equation}\label{idf}
IDF(w_l)=ln\Big(\frac{N-n(w_l)+0.5}{n(w_l)+0.5}+1\Big),
\end{equation}
where $N$ represents the total number of documents in the indexed KG, while $n(w_l)$ signifies the count of documents containing the word $w_l$.

For instances where the cell mention corresponds to a number or a date, it is inappropriate to link it to the KG. In such situations, we assign a linking score of 0 to the cell. This strategy enables us to address this specific issue within the context of the deep learning prediction model. Subsequently, the deep learning model addresses this challenge by incorporating the column type representation generation task, which will be further discussed in the subsequent subsection.

Following the completion of the linking process, we acquire a KG entity set $E_{m_c^r}$ for each cell mention $m_c^r$.
\\

\begin{figure}[t]
  \centering
  \includegraphics[width=0.9\linewidth]{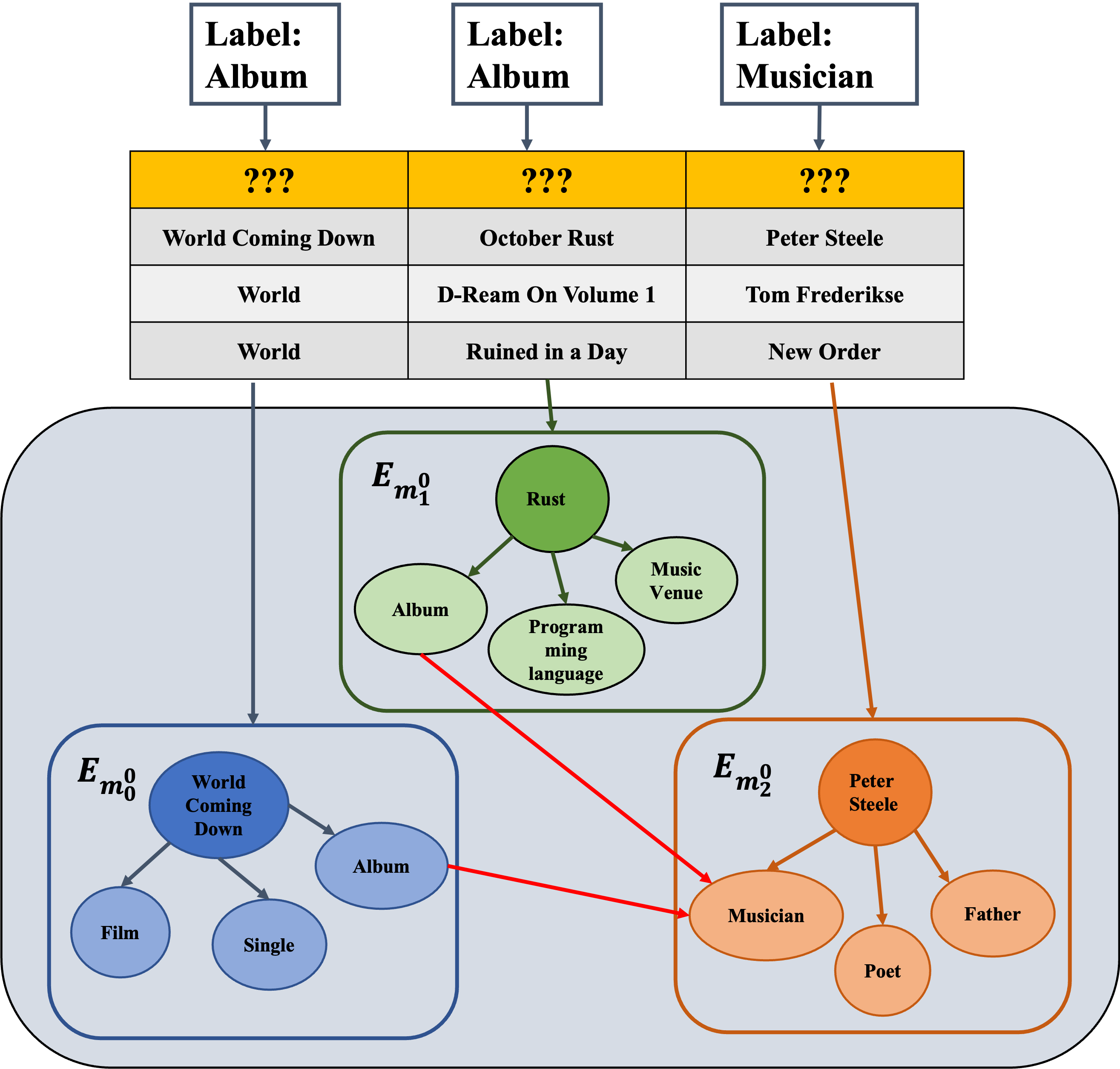}
  \caption{An example from the SemTab \cite{hassanzadeh_oktie_2019_3518539} dataset, showcasing three KG entity sets: $E_{m_0^0}$, $E_{m_1^0}$, and $E_{m_2^0}$. The figure includes a red link representing a connection between entities in the one-hop neighbor. For instance, the entity \textit{Rust}, which corresponds to an album, has a one-hop neighbor entity: \textit{Peter Steele}, representing a musician. This connection suggests a higher likelihood that these two entities are the correct representation for the cell mentions \textit{Rust} and \textit{Peter Steele}.}
  \label{fig:overlap_fig}
  \vspace{-0.3cm}
\end{figure}

\noindent \textbf{Step 2: Filters on rows and entities:} 
Upon completion of step 1, we proceed with filtering the extracted KG entities. The goal of this step is to create a pruned set $\hat{E_{m_c^r}}$ from the KG entity set $E_{m_c^r}$ for each cell mention $m_c^r$, and to filter the table rows with the linking score. As illustrated in \autoref{fig:overlap_fig}, cell mentions originating from the same row but occupying different columns in a table can demonstrate close relationships, suggesting a potential connection between their corresponding KG entities \cite{2021_mtab4wikidata,oliveira2019adog,abdelmageed2020jentab}.

To capture this relationship, we define the overlapping entity set $\hat{E_{m_c^r}}$ for a cell mention $m_c^r$ in row $r$ and column $c$. This is achieved by intersecting its entity set with the one-hop neighbor entity set of all other columns. This intersecting approach also helps filter out noise introduced by the one-hop neighbor entity set. Mathematically, this can be expressed as demonstrated in \autoref{overlap_entity}:
\vspace{-0.1cm}
\begin{equation}\label{overlap_entity}
	\hat{E_{m_{c1}^r}}=\bigcup_{c1\neq c2}^C \Big({E_{m_{c1}^r}\ \cap \mathcal N(E_{m_{c2}^r})\Big)},
\end{equation}
where $C$ represents the column set of table $T$ and $c1, c2$ represent columns in set $C$ of table $T$, $\mathcal N(E_{m_{c2}^r})$ represents  the one hop neighbor set for $N(E_{m_{c2}^r})$. Following \autoref{overlap_entity}, we consider the set $\hat{E_{m_{c1}^r}}$ as the candidate entity set for the cell $m_{c1}^r$. This approach incorporates inter column information, helping to prevent errors that may have occurred in step 1 from propagating.

In the process of filtering the table rows, we evaluate the linking score of each cell. After completing the linking procedure, for every candidate entity $e$ associated with the cell mention $m_c^r$ in row $c$, we obtain its BM25 linking score $ls_e$. As presented in \autoref{max_linking_score}, we determine the cell's linking score $ls_{m_c^r}$ for the cell mention $m_c^r$ by selecting the maximum BM25 linking score among its candidate entities.

\vspace{-0.3cm}
\begin{equation}\label{max_linking_score}
	ls_{m_c^r}= max(ls_e), \ \ (e\in \hat{E_{m_c^r}}).
\end{equation}

After obtaining the linking score for each cell in a row, the process of selecting the top-$k$ rows for prediction involves considering the linking score of a row $r$ as the sum of all its cells' linking scores:
\vspace{-0.3cm}
\begin{equation}\label{row_linking_score}
	ls_{r}= \sum_c^C{ls_{m_c^r}}.
\end{equation}

A higher linking score for a row indicates greater reliability. Rows with elevated scores are less likely to have incorrect or missing links in the KG. This reliability is due to the typical pre-training of deep learning LMs with KG information. Consequently, the top-$k$ rows chosen based on these high scores are considered more appropriate for predictions by deep learning LMs. During each iteration of the loop, we insert rows into an empty table. These rows are sorted in descending order based on the sum of the linking scores of their cell mentions. Once all iterations are complete, we retain only the top-$k$ rows from this new table, effectively creating the filtered table.

To evaluate the credibility of a candidate entity, we introduce the concept of an overlapping score $os_e$ for each entity $e$ in the candidate entity set $\hat{E_{m_c^r}}$ associated with the cell $m_c^r$ in row $r$. The overlapping score, defined by \autoref{overlapping_score}, quantifies how frequently the candidate entity $e$ appears in the one-hop neighbor set of candidate entities from other columns. A higher overlapping score signifies a greater level of reliability for the entity $e$.

\vspace{-0.5cm}
\begin{equation}\label{overlapping_score}
	os_e= \Big|\ e\ \bowtie \bigcup_{c1\neq c2}^C \mathcal N(E_{m_{c2}^r})\ \Big|\ ,\ \ (e\in \hat{E_{m_{c1}^r}}).
\end{equation}

\noindent \textbf{Step 3: Candidate type generation:} 
In this step, we determine potential types for the columns in the table and introduce a novel overlapping score to measure the intersection between entity nodes in the KG. 
Given a table $T$ and a target column $c$ in $T$'s column set $C$, a set of candidate types $CT=\{ct_0,ct_1,...,ct_j\}$ are the label of type entities in the KG, which could be the semantic type that describes column $c$. In this step, we aim to generate candidate types $ct$ for every target column $c$ in table $T$.

In KGLink, we consider the candidate types from the candidate type set $CT$ for each column to assist with the column annotation task. It's worth noting that these candidate types may not always exist in the type hierarchy of the types in set $D_{type}$, as illustrated in Figure \autoref{fig:tg_issue}. For example, even though \textit{Name} is a type extracted from the KG, it does not exist in the type hierarchy of \textit{Basketball player} or \textit{Athlete}.

While we can directly extract the semantic type of an entity from the KG, the default granularity of KG's types often falls short of our needs. For instance, consider mention \textit{Peter Steele} in \autoref{fig:overlap_fig}, which ideally should be categorized as \textit{Musician}. However, in the KG, \textit{Peter Steele} is labeled as \textit{Human}, even though \textit{Musician} is present as an entity in the one-hop neighbor of \textit{Peter Steele} within the KG. To overcome this limitation, our objective is to generate more suitable candidate types that better align with our specific context and requirements, as depicted in \autoref{fig:overlap_fig}.

In our methodology, we take advantage of the observation that type entities frequently emerge within the one-hop neighborhood of candidate entities. For every candidate entity $e$ within column $c$ from the pruned entity set $\hat{E_{m_c^*}}$, our process unfolds as follows:

\vspace{-0.5cm}
\begin{equation}
\hat{E_{m_c^*}}=\bigcup_{r}^R \hat{E_{m_c^r}},\ (c\in C),
\end{equation}
where, $\hat{E_{m_c^*}}$ symbolizes the union set of all pruned entity sets gathered from each row within column $c$, while $R$ stands for the set of rows in table $T$. For each candidate entity $e$ in $\hat{E_{m_c^*}}$, we attribute its overlapping score $os_e$ to all entities within its one-hop neighborhood. As a one-hop neighbor entity could have multiple connections to entities in $\hat{E_{m_c^*}}$, we compute the candidate type score $cts$ by summing up all overlapping scores of its neighbors:

\vspace{-0.6cm}
\begin{align}\label{cts_score}
\begin{split} 
	cts_{ct_c}&= \sum_{r2\neq r1}^R |ct_c\cap \ \mathcal N(e^{r2})|\times os_{e^{r2}}, \\ 
	&(e^{r2} \in \hat{E_{m_c^{r2}}},\ ct_c\in \ \mathcal N(e^{r1})),
\end{split}
\end{align}
where $R$ signifies the set of rows within table $T$, and $r1, r2$ represent specific rows within this set. Moreover, $ct_c$ denotes the candidate type of column $c$.

It's noteworthy that during the selection of $ct_c$ from the set $\mathcal N(e^{r1})$, we apply a label-based filter. This filter ensures the exclusion of entities classified under the named entity schema categories \textit{PERSON} or \textit{DATE}. The rationale behind this is that entities belonging to these types are not well-suited to represent column types within a table. 

As a result, the ultimate candidate type is more likely to be chosen from entities that are interconnected with higher-reliability entities within the same column. This selection criterion significantly increases the likelihood of identifying the most appropriate candidate type.

After obtaining the candidate types, we concatenate them at the top of each column to form the processed table. Additionally, during the training preparation phase, the mask token or the ground truth label is also concatenated before the candidate types, creating the masked table and the ground truth table, respectively, in order to facilitate the column type representation generation task. It is crucial to note that if our algorithm returns no candidate types, the corresponding space for candidate types is replaced with padding tokens. For numeric columns, the candidate types are replaced with the column's mean, variance, and average value.

Meanwhile, to set the stage for feature vector creation, we begin by constructing the feature sequence. Within the filtered table, we proceed to select the first cell from each predicted column. This selection is based on the filter from the last step, specifically opting for the cell with the best total linking score. Subsequently, we identify the KG entity with the highest linking score for that cell. Next, we serialize entity $e$ and all its one hop neighbor, to prepare the feature sequence $S(e)$ to let it ready for encoding in the deep learning part of our model:
\begin{equation}
    S(e)=s||\Big(\Big\Vert_{o \in \mathcal N(e)}p||o\Big),
\end{equation}
where $p$ represents the predicate connecting entity $e$ with another entity in $ N(e)$, $||$ signifies the concatenation operation. Note that when no entities are retrieved from the KG (e.g., in the case of a numeric column or when there are no matched entities for the cell mention in the KG), we set $S(e)$ to a padding sequence, which only contains the padding token.

\subsection{Part 2: Deep learning based model}\label{tdl}

In this section, we integrate table context with a deep learning-based pre-trained language model (PLM) using multi-task learning. This integration serves to address the type granularity issue and to enhance the overall model performance. The section is organized into three distinct steps, each designed with a specific objective. Through the integration of these steps, we harness the power of table context, bolster the model's capabilities, and ultimately achieve enhanced performance and effectiveness.
\\




\noindent \textbf{Step 1: Table serialization:} In this section, we proceed with serializing the table into a sequence format to facilitate multi-class classification. In KGLink, we adapted the multi-column prediction serialization method from Doduo \cite{suhara2022annotating} to align it with the requirements of the column type representation generation task.

Given that PLMs operate on tokenized sequences, it's essential to serialize tables into token sequences to fully leverage the structural advantages of PLMs. In certain prediction models that focus on single columns, such as Sherlock \cite{hulsebos2019sherlock}, table serialization occurred on a column basis. This implied concatenating cells from the same column together. For instance, in the case of column $c_i$, within single-column prediction models, the serialization would appear as follows:
\begin{equation}
S(c_i)_{single}::=[CLS]\ {tok}_i^1\ {tok}_i^2\ ...\ {tok}_i^m\ [SEP].
\end{equation}

In this context, it's important to note that [CLS] and [SEP] tokens are special tokens within the BERT model, signifying the start and end of a sequence, respectively. The ${tok}$ variable represents the tokens that result from the sequence's word tokenization. These tokens become instrumental in generating representation vectors through PLMs.

However, single-column prediction models lack the advantage of contextual information from other columns within the same table. This shortcoming arises because these models independently predict each table. Consequently, they overlook the inter-column relationships that hold significance for tasks like column annotation \cite{chen2019learning,khurana2021semantic,zhang2020sato}.

To holistically embrace the contextual relationships among columns within a table, we employ a serialization approach introduced by Doduo \cite{suhara2022annotating}, which is a multi-column prediction model. This serialization method treats the entire table as a sequence, focusing on serialization by column:
\vspace{-0.5cm}

\begin{align}
\begin{split}
S(T)_{multi}::=&[CLS]\ {tok}_1^1\ ...\ {tok}_1^m\ [CLS]\ {tok}_2^1...\\  \ &...[CLS]\ {tok}_n^1\ ...\ {tok}_n^m\ [SEP].
\end{split}
\end{align}

In contrast to the serialization technique employed by single-column prediction models, this strategy involves incorporating the $[CLS]$ token at the commencement of each column. This step facilitates the serialization of the complete table into a single sequence. Consequently, the PLM can effectively grasp the contextual details from the entire table. During downstream tasks, the representation of the $[CLS]$ token from each column can be passed to the output layer following the final PLM layer.

blueAs for KGLink's deep learning aspect, to align with most of our baselines, we undertook the fine-tuning of the BERT \cite{devlin2018bert} model:
\begin{equation}
    Y=BERT(S),
\end{equation}
where $S$ represents the input sequence, $Y\in \mathbb{R}^d$ represents the encoding vector for the input sequence $S$. Note that the BERT PLM can be substituted by other more powerful language models with encoder structure to further improve the performance.
\\

\begin{figure}
  \centering
\includegraphics[width=\linewidth]{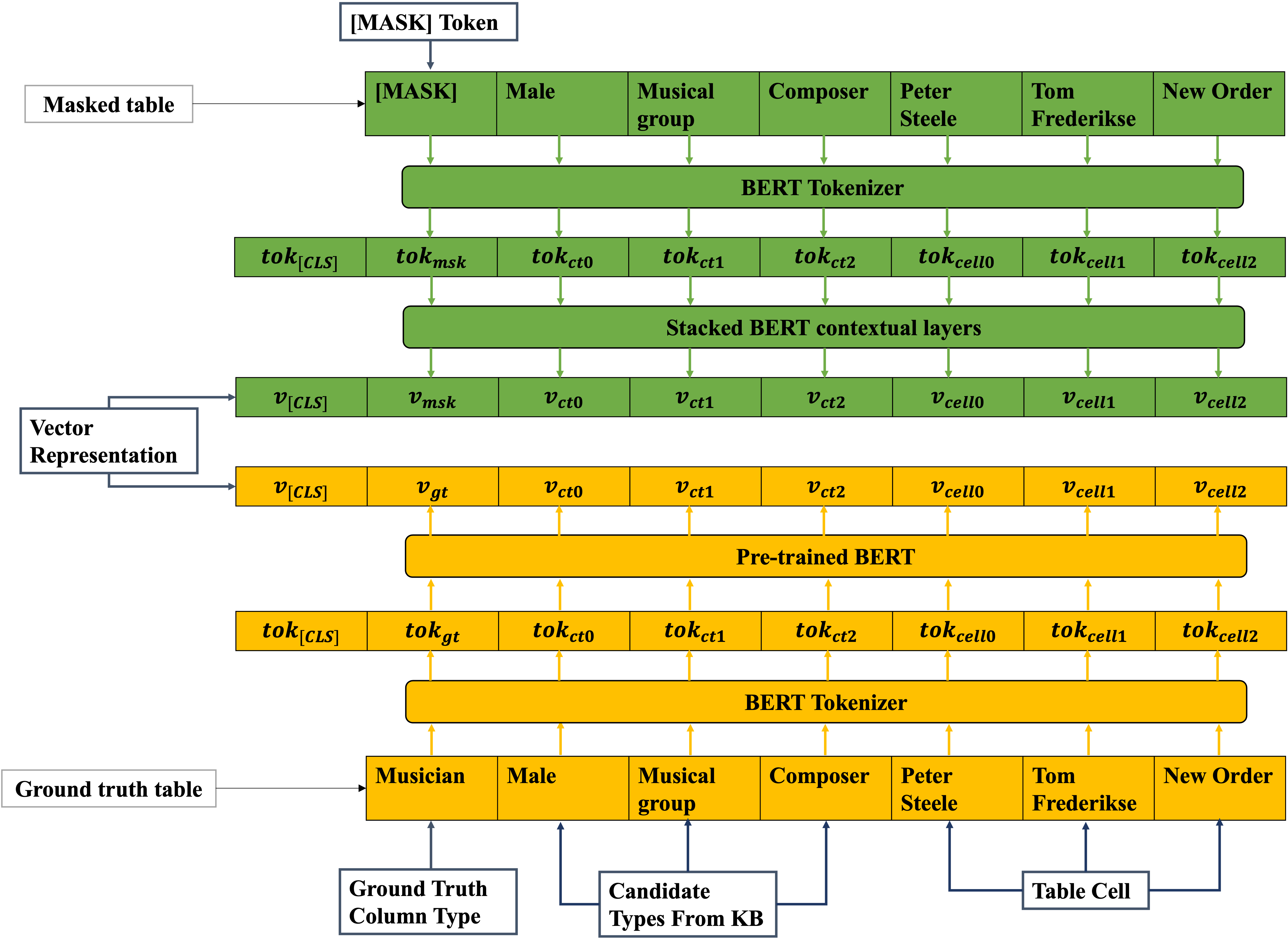}
  \caption{An example from the SemTab \cite{hassanzadeh_oktie_2019_3518539} dataset, the column data would be tokenized and then encoded using pre-trained BERT, after that, the generated $v_{msk}$ and $v_{gt}$ vector would be used in DMLM \cite{wang2020learning} loss to let the model learn the column type representation.}
  \vspace{-0.2cm}
  \label{fig:dmlm}
\end{figure}

\noindent \textbf{Step 2: Column-type representation generation:} At this step, we introduce the column-type representation generation task, which is centered around extracting the underlying semantic information of the column type by restoring the ground truth candidate type representation from masked tokens.

To implement this task within KGLink, we establish the column-type representation generation task as a sub-task. Here, we leverage the BERT model's \cite{devlin2018bert} generation capabilities.

To integrate contextual information, we generate both a ground truth table and a masked table during training. This allows the model to learn how to recover the masked token with supervision from the ground truth table. It's important to note that during model evaluation, the ground truth table is not created to prevent leakage.

For a given table $T$, which contains target columns $c$ in it's column set $C$, the column's ground truth label $l$, the ground truth table $T_{gt}$ is formed by concatenating each target column $c$ from $T$ with its corresponding $l$ at the top of the column. Conversely, the masked table $T_{msk}$ is formed by concatenating each column $c$ from $T$ with the mask token provided by the language model.

As depicted in \autoref{fig:dmlm}, we prefix the ground truth label of the column type, such as \textit{Musician}, and the set of generated candidate types from the knowledge graph (KG), such as \textit{Male}, \textit{Musical group}, and \textit{Composer}, to the beginning of each serialized column in the table. This modified table is referred to as the \textit{ground truth table}. Furthermore, we utilize the [MASK] token provided by the BERT model to replace the actual ground truth label in the \textit{ground truth table}, resulting in the creation of the \textit{masked table}.

Throughout the training process, the goal is to reconstruct the original ground truth label from the [MASK] token in the \textit{masked table}, while accounting for the contextual details from the table and the extracted candidate types. It's important to underscore that the ground truth label in the \textit{ground truth table} is also embedded with the table's context. Consequently, optimizing the retrieved representation of the [MASK] token from the masked table takes this contextual information into consideration.

To accomplish this, we employ the DMLM (Distilled Masked Language Model) loss \cite{wang2020learning} in the process of recovering the [MASK] token. This loss function serves to enhance the accuracy of predicting the original ground truth label.

\vspace{-0.3cm}
\begin{equation}\label{dmlm_loss}
	\mathcal{L}_{DMLM}(Y_{msk},Y_{gt})=-\sum_{voc}^V y_{msk}^{voc}\log y_{gt}^{voc},
\end{equation}

\vspace{-0.3cm}
\begin{equation}\label{dmlm_loss_softmax}
	Y=Softmax(W_o(\frac{H}{\mathcal{T}})),
\end{equation}
where $V$ is the vocabulary size, $H$ is the generated vector representation, $W_o$ is a matrix that projects the output vector to the vocabulary space, and $voc$ is a variable iterate through the number of vocabulary words. The $Y_{msk}$ is the representation in the vocabulary space generated from the [MASK] token, and the $Y_{gt}$ is the ground truth label's vector representation in the vocabulary space. The $y_{msk}^{voc}$ and $y_{gt}^{voc}$ are the weight of the corresponding word in the vocabulary space. The $\mathcal{T}$ is a scale factor on the output embedding vector. 

By incorporating the DMLM loss, the model is trained to generate the column type representation from the [MASK] token, aiming to match the ground truth label vector as closely as possible. This approach is beneficial even for annotating numeric columns, as it guides the model to generate accurate column type representations and aids in the column type annotation process.
\\

\noindent \textbf{Step 3: The adaptive combined loss:} 
In this section, we combine the column representation vector with the feature vector and apply the Uncertainty Weights (UW) method proposed by Kendall et al. \cite{kendall2018multi} to optimize the column-type annotation and column-type representation generation tasks simultaneously.

Denote the representation of the [CLS] token as the column representation vector $Y_{cls}$ and the feature vector as $Y_{fv}$. We obtain the combined vector $Y_{col}$ as follows:
\begin{equation}
    Y_{col}=\phi(Y_{cls},Y_{fv}),
\end{equation}
where $\phi(\cdot):\mathbb{R}^d \times \mathbb{R}^d \rightarrow \mathbb{R}^d$ is a composition function.

Considering that the column type annotation task involves a multi-class classification problem, we utilize the Cross-Entropy loss for this task: 
\vspace{-0.2cm}
\begin{equation}\label{cross_entropy}
	\mathcal{L}_{CrossEntropy}(Y'_{col},L)=-\log\frac{e^{y'_{gt}}}{\sum_{i=1}^{|L|} e^{y'_{i}}},
\end{equation}
where $Y'_{col}$ is the combined vector in the classification space that has dimension $|L|$, $y'_{gt}$ is the weight that the model predicts out the ground truth label as the true label, $y'_{i}$ is the weight that the model predicts out the $i-th$ label as the true label.


To seamlessly merge this task with the column-type representation generation task, we must dynamically combine the DMLM loss and the Cross-Entropy loss to cater to different datasets. Thus, we substitute the loss of tasks in \cite{kendall2018multi}  by DMLM and Cross Entropy loss:
\begin{align}
\begin{split}\label{uw}
	\mathcal{L}_{total}(H,\sigma_0,\sigma_1)=&\frac{1}{2\sigma_0^2}\mathcal{L}_{DMLM}+\frac{1}{2\sigma_1^2}\mathcal{L}_{CrossEntropy}\\&+\log\sigma_0\sigma_1,
\end{split}
\end{align}
where $\sigma_0$ and $\sigma_1$ are tunable variables.
\subsection{Complexity} \label{complexity}

In the KGLink, we first link cell mentions to the KG, then calculate the overlapping score and perform entity filtering. After that, we calculate the candidate type score and filter the candidate type. Next, we serialize the table and utilize the PLM to encode tables and perform the classification task and the column type representation generation subtask. 

\begin{figure}
  \centering
\includegraphics[width=\linewidth]{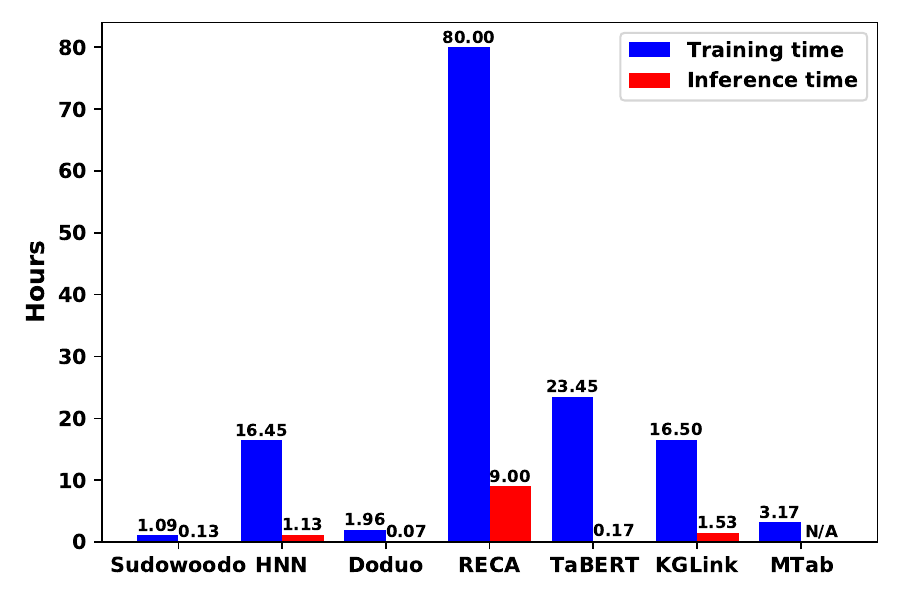}
\vspace{-0.8cm}
  \caption{The time chart of KGLink and other baselines on VizNet}
  \vspace{-0.5cm}
  \label{fig:runtime}
\end{figure}

For a dataset, considering the maximum number of columns $C$, rows $R$, tables $T$, epochs $P$, the size of the label type set $D$, the vocabulary size $V$, the maximum input sequence $B$, and the embedding dimension of the PLM $d$, and the maximum number of KG retrieved entities as $E$, and their maximum one hop neighbor number as $N$, we calculate the time and space complexity as follow:

\noindent \textbf{Time Complexity:} 
The KGLink traverses each column and row to link cells to the KG, obtaining the linking score and candidate entity set, which takes $O(RC)$ time. The candidate type score calculation and the candidate type filtering take $O(RC^2E^2T)$ time. Calculating the candidate type score requires $O(RCNE)$. Thus, the data pre-processing time complexity of KGLink is $O(RC)+O(RC^2E^2T)+O(RCNE)=O(RCE(N+CET))$. Since we serialize the input table, encoding tables and feature vectors would take $O(BdP(T+TC)(B+d))$ time \cite{devlin2018bert, vaswani2017attention}. The classification task would take $O(d^2DCP)$, and the column type representation generation subtask would take $O(VDCP)$. Hence the overall time complexity of KGLink is: $O(RCE(N+CET)+BdP(T+TC)(B+d)+d^2DCP+VDCP)=O(RCE(N+CET)+BdP(T+TC)(B+d)+(d^2+V)DCP)$ The KGLink is linear in time complexity. Compared with the state-of-the-art method RECA, which has exponential complexity to tables. KGLink has better scalability for the growing size of datasets. The runtime chart of KGLink and other baselines are provided in \autoref{fig:runtime}; we only show the runtimes on the VizNet dataset due to the space limitation.

\noindent \textbf{Space Complexity:} The KGLink requires $O(RCE)$ space to store each table and their corresponding entities from the KG. The linking score needs $O(E)$ to store. Calculating the overlapping score requires $O(RCE+E)$ to store the overlapping score for all entities, and calculating the candidate type score requires $O(RCNE+E)$ to score the candidate type for all entities within one-hop neighbor of the KG-retrieved entities. Preparing input and encoding columns requires $O(T+d(B+d))$ space since we serialize each table into one input sequence. It is $O(Bd)$ space for the embeddings, $O(d^2)$ for self-attention matrices, and $O(d(V+D))$ for the classification step and combined adaptive loss calculation. Hence, the overall space complexity is: $max(O(RCNE),O(T+d(B+d),O(Bd),O(d^2),O(d(V+D))$.

\section{Experiment Results}
We compared KGLink with six state-of-the-art models: MTab \cite{2021_mtab4wikidata}, TaBERT \cite{yin2020tabert}, Doduo \cite{suhara2022annotating}, HNN \cite{chen2019learning}, Sudowoodo \cite{wang2022sudowoodo}, and RECA \cite{sun2023reca}. 

In our experiment, if we cannot find any linkages in the KG for a specific mention, we assign a linking score of 0 to that mention.

For each cell mention, we determine whether it represents a number or a date with the named entity schema. If it does, it is unsuitable for linking to the KG. In such cases, we set the linking score of that cell to 0, allowing the deep learning prediction model to address this issue through the column-type representation generation task.

\subsection{Dataset} We evaluated our approach using two benchmark datasets. The first dataset, SemTab \cite{hassanzadeh_oktie_2019_3518539}, consists of 3,048 tables and 7,587 columns, with 69 rows and 4.5 columns per table on average. It is derived from Wikipedia, DBpedia, and T2Dv2 Gold Standard data sources. We focused on rounds 1, 3, and 4 from the 2019 version of SemTab, similar to the approach taken by RECA \cite{sun2023reca}. The SemTab dataset annotates 275 column types corresponding to DBpedia KG entities. We excluded round 2 from our evaluation due to columns with multiple ground truth labels, which was unsuitable for our comparison with VizNet.

The second dataset is a modified version of VizNet \cite{hu2019viznet}, introduced by Zhang et al. \cite{zhang2020sato}. This version is a subset of the original VizNet corpus \cite{hu2019viznet} specifically extracted for the column type annotation task, consisting of 78,733 web tables with 119,360 columns annotated across 78 column types, with 20 rows and 2.3 columns per table on average. Following the methodology of RECA \cite{sun2023reca}, we utilized the multi-column sub-dataset, which includes 32,265 web tables and 73,034 columns annotated with 77 column types. It's important to note that each web table in this dataset contains multiple columns. Notably, the SemTab dataset offers more detailed column type annotations compared to the modified VizNet dataset, but it is smaller in size.

\begin{table}[t!]
  \caption{KGLink performance on the SemTab dataset and the VizNet dataset}
  \label{tab:performance}
  \centering
  \scalebox{0.95}{
  \begin{tabular}{c|cc|cc}
    \toprule
    \multirow{2}{*}{Model}&
	\multicolumn{2}{c}{SemTab}&
	\multicolumn{2}{c}{VizNet} \\
	\cline{2-5}
  & Accuracy & Weighted F1  & Accuracy & Weighted F1 \\
    \midrule
     MTab \cite{2021_mtab4wikidata} &  \textbf{89.10} &  -
     &  38.21 &  - \\
    TaBERT \cite{yin2020tabert} &  72.69& 71.21   & 94.68  & 94.07 \\
    Doduo \cite{suhara2022annotating} & 84.06 &	82.43 
    & 95.40& 95.06 \\
    HNN \cite{chen2019learning} & 66.54& 65.12  
    & 66.89 & 68.82\\
    Sudowoodo \cite{wang2022sudowoodo} & 79.34 & 79.24  
    & 91.57 & 91.08  \\
    RECA \cite{sun2023reca} & 86.12 & 84.91 
    & 93.25 & 93.18  \\
    \textbf{KGLink} & 87.12 & \textbf{85.78}	
    & \textbf{96.28}& \textbf{96.07} \\
    \bottomrule
  \end{tabular}}
  \vspace{-0.3cm}
\end{table}

\subsection{Experimental Settings}

\begin{table*}[t!]
  \caption{The ablation study result of KGLink}
  \label{tab:abl_performance}
  \centering
  \scalebox{0.95}{
  \begin{tabular}{c|cccc|cc|cc}
    \toprule
    \multirow{2}{*}{Model}&
    \multirow{2}{*}{PLM}&
    \multirow{2}{*}{Multi-task}&
    \multirow{2}{*}{KG candidate type}&
    \multirow{2}{*}{KG feature vector}&
	\multicolumn{2}{c}{SemTab}&
	\multicolumn{2}{c}{VizNet} \\
	\cline{6-9}
  & &  & & & Accuracy & Weighted F1  & Accuracy & Weighted F1 \\
    \midrule
     KGLink $w/o	\ msk$ & BERT & & $\checkmark$ & $\checkmark$  & 86.14 &  84.54
     & 95.95 & 95.67\\
    KGLink $w/o	\ ct$ & BERT & $\checkmark$ & &  & 86.27& 84.56
    & 95.83& 95.48\\
    KGLink $w/o	\ fv$ & BERT & $\checkmark$ & & $\checkmark$  & 87.02 & 85.68 
    & 95.98& 95.70 \\
     KGLink DeBERTa &  DeBERTa &  $\checkmark$ & $\checkmark$ & $\checkmark$  &  \textbf{87.24} &  \textbf{85.81}
    &  \textbf{96.98} &  \textbf{96.37} \\
    \textbf{KGLink} & BERT&  $\checkmark$ & $\checkmark$ & $\checkmark$  & 87.12 & 85.78	
    & 96.28& 96.07\\
    \bottomrule
  \end{tabular}}
  \vspace{-0.5cm}
\end{table*}

We used the AdamW optimizer with $\epsilon=10^{-6}$ and an initial learning rate of $3\times 10^{-5}$. The learning rate was linearly decayed without warm-up. The scale factor $\mathcal{T}>1$ is set to 2, as suggested by Hinton et al. \cite{hinton2015distilling}. KGLink was trained for 50 epochs on the SemTab dataset, and 20 epochs on the VizNet dataset, we applied a 0.1 dropout rate on the SemTab dataset, while 0.2 on the VizNet dataset since it contains more training tables, the batch size was set to 16 on both datasets, and early stopping was also applied. The experimental settings for TaBERT and Doduo were the same as KGLink. For HNN and RECA, we use the same experiment setting as their original paper \cite{chen2019learning,sun2023reca}. Despite Sudowoodo's semi-supervised and unsupervised ability, we utilize the same amount of training data with other baselines, making it a full-supervised model, which includes all labeled columns in the training set. For MTab, due to the difference in problem definition, we did not include the weighted F1 score. We translate the label on VizNet dataset to WikiData KG entities to make MTab work on VizNet dataset.
We run each experiment 3 times and calculate the average metric in \autoref{tab:performance}.

The training, validation, and testing set proportions were set at 7:1:2 for all experiments, and we maintained the original sample proportion of each class in all splits. The models were implemented using PyTorch and the Transformer library. We utilized the open-source official implementations of the baselines, and the experiments were conducted on an NVIDIA Tesla V100 GPU. Evaluation metrics included accuracy and weighted F1 scores for both datasets.

We used Elasticsearch, a distributed search and analytics engine, to index the WikiData KG and generate the BM25 entity linking scores for the KG entity callback. We retrieved up to 10 entities from the KG for each cell mention. From the KG entities in each column, we generated up to 3 candidate types per row. The label-based filter used the spacy NLP tool to detect the named entity schema and filter entities labeled as \textit{PERSON} or \textit{DATE}.

Given that BERT has an input sequence length limit of 512, we restrict the token length per column in a table to 64. Additionally, we impose a maximum limit of 8 columns per table. If a table contains more than 8 columns, we divide it into multiple tables after the candidate type generation step and conduct the encoding and annotation process separately.

\subsection{Main Results}
\autoref{tab:performance} presents accuracy and weighted F1 scores for the column-type annotation task on the SemTab and modified VizNet datasets, showcasing KGLink's superior or comparable performance to the baseline models.

Across both datasets, KGLink consistently outperforms other baselines, enhancing accuracy and weighted F1 scores, highlighting the effectiveness of incorporating KG information and employing a multi-task learning approach.

In the VizNet dataset, KGLink outperforms TaBERT, Doduo, and Sudowoodo by leveraging KG information for improved predictions. Notably, despite being designed based on Doduo's table serialization method and benefiting from KG-extracted information, KGLink surpasses Doduo across all metrics.




On both datasets, KGLink outperforms HNN across all evaluation metrics due to its proficient use of knowledge within the BERT PLM and its refined integration with the KG. Although HNN also integrates KG information, it appears to be more vulnerable to noise from the entity linking process and does not fully capitalize on the advantages of KG information.

Regarding RECA, its ability to capture inter-table information makes it less sensitive to training data distribution, achieving the current state-of-the-art performance. However, since RECA ignores intra-table information and does not integrate KG information, KGLink maintains a performance edge across these metrics on the VizNet dataset. On the SemTab dataset, 
RECA achieves sub-optimal results because of the dataset's nature. The SemTab only consists of 3048 tables but 275 column type classes. This also shows that the approach introducing extra information (KGLink) or the approach better utilizing the existing information (RECA) would be less affected when the data amount is constrained.



Regarding Doduo, on the VizNet dataset, the accuracy of KGLink increased by 0.92\%, and the weighted F1 increased by 1.06\%. 
Since Doduo and KGLink apply the same table serialization method and the same PLM (BERT), both methods can benefit from capturing intra-table information and the prior knowledge of the PLM, hence achieving competitive performance on the VizNet dataset. According to \autoref{tab:no_ct}, KGLink still outperforms Doduo on columns with no linkages to KG because of the application of the multi-task learning component, and the model's better ability to process numeric columns.

Regarding MTab, since the labels of the SemTab dataset are extracted from KG, MTab achieves the highest accuracy score on the SemTab dataset. However, on the VizNet dataset, the MTab achieves the worst accuracy, it is because the VizNet dataset consists of 12.8\% numeric columns, which is hard for solely KG based method prediction. Furthermore, the label of the VizNet dataset is not KG entities, although we translated them to KG entities, this still limit the performance of MTab.

\vspace{-0.2cm}

\section{Analysis}
\vspace{-0.2cm}
\subsection{Ablation Study}

To demonstrate the impact of our components, we conducted tests on variations of the KGLink model. These variants are referred to as 
KGLink $w/o\ msk$, KGLink $w/o\ ct$, KGLink $w/o\ fv$, and KGLink DeBERTa. The results of these ablation studies are displayed in \autoref{tab:abl_performance}. 

The KGLink $w/o\ msk$ model excludes the generation of representations from the [MASK] token. The KGLink $w/o\ ct$ model excludes all KG information (the candidate types and the feature vector). The KGLink $w/o\ fv$ removes the feature vector generated from KG information. The KGLink DeBERTa substitutes the BERT PLM with a more advanced transformer based PLM DeBERTa \cite{he2020deberta}. Each variant allows us to isolate and assess the individual contributions of the corresponding components to the overall model performance.


KGLink outperforms the KGLink $w/o\ msk$ model on the VizNet dataset. This could be attributed to the VizNet dataset's coarser label granularity than the SemTab dataset, making the column type representation generation subtask less significant. Additionally, as shown in \autoref{tab:dataset_nolink}, the VizNet dataset contains numeric columns that do not exist in the SemTab dataset. These columns are not connected to KG and, therefore, cannot benefit from KG information. 

KGLink outperforms the KGLink $w/o\ ct$ model on both datasets, underscoring the significance of KG-extracted information. 



KGLink also outperforms the KGLink $w/o\ fv$ model on both datasets. According to \autoref{tab:dataset_nolink}, the feature vector could be considered a supplement when all KG information is filtered out and generates no candidate types. This makes the feature vector able to enhance the scalability of KGLink further.

The KGLink DeBERTa model outperforms the KGLink model, this demonstrate that more powerful PLM encoder are likely to bring further improvements to KGLink. 




The differing increases in the two metrics from KGLinlk $w/o\ ct$ to KGLink $w/o\ fv$ and from KGLink $w/o\ fv$ to KGLink also illustrate that carefully selected candidate types are more important than KG information in a feature vector, despite the former having worse coverage across tables (Indicated in \autoref{tab:dataset_nolink}).

\subsection{Parameter sensitivity}
\begin{table}[t]
\centering
\caption{The link statistics between datasets and the WikiData KG}
\label{tab:dataset_nolink}
\begin{tabular}{l|l|l}
\hline
                                     & \textbf{SemTab}       & \textbf{VizNet}        \\ \hline
\multicolumn{1}{l|}{Numeric columns} & 0 (0\%)      & 9489 (12.8\%) \\ \hline
\multicolumn{1}{l|}{Non-numeric columns $w/o\ fv$} & 0 (0\%)       & 9278 (12.5\%)  \\ \hline
\multicolumn{1}{l|}{Non-numeric columns $w/o\ ct$}             & 1144 (15.1\%) & 55374 (74.7\%) \\ \hline
\multicolumn{1}{l|}{Total columns}   & 7587 (100\%) & 74141 (100\%) \\ \hline
\end{tabular}
\vspace{-0.5cm}
\end{table}

\autoref{tab:dataset_nolink} illustrates the connection scenarios of KGLink on two datasets. We segregated this test into numeric columns and non-numeric columns. If all cells from a column are numeric, we classify this column as numeric; otherwise, it is a non-numeric column.

Non-numeric columns $w/o\ fv$ stands for non-numeric columns with no KG-extracted information to generate feature vectors, meaning it has zero linkage to the KG. Non-numeric columns $w/o\ ct$ stands for non-numeric columns with no candidate type extracted by our method from the KG. This could be because there is no linkage for the column at all, or all entities retrieved from the KG were filtered by our method. These non-numeric columns are typically hard to annotate and involve cells with extended sentences, such as a lengthy address in a column labeled \textit{address}, or short abbreviation codes (e.g., In a column with the type \textit{position}, the abbreviation code \textit{PF} in the column cell might stand for \textit{Power Forward}, a specific position in a basketball game.). 

Since the tables from SemTab were derived from the KG, only $15.1\%$ of columns have no candidate type extracted from the KG by our method. Moreover, no column lacks entities extracted from the KG to generate feature vectors. For the VizNet dataset, even though $74.7\%$ of columns have no candidate type extracted, after applying the feature vector, only $12.5\%$ of non-numeric columns lack KG information. 

In \autoref{tab:no_ct}, we also compare the prediction performance of KGLink with other baselines when the columns cannot connect to KG. We select columns from the modified VizNet dataset whose entire table has no linkage to the KG from the original test set. We adopt this strategy because if other columns from the same table have linkages, the column with no linkage would also partially benefit from the KG information of other columns. This subset comprises 315 tables and 612 columns in total, with 26 different labels. Of these columns, 556 are numeric, and 56 are non-numeric. We calculated their accuracy separately.

The results demonstrate that KGLink can maintain competitive effectiveness even when it cannot find related KG information for columns. Since all baselines other than HNN applied the PLM to encode the table, the results also underscore the crucial role of prior knowledge from the PLM in both numeric and non-numeric column annotation. Notably, KGLink, Doduo, and TaBERT outperform RECA and Sudowoodo, both of which are PLM-based baselines, particularly in non-numeric columns. This superior performance is attributed to KGLink, Doduo, and TaBERT's ability to capture intra-table information between columns. The observed phenomenon underscores the importance of intra-table information in the column type annotation task.
\begin{figure}[t]
 \centering
  \subfigure[\footnotesize{Sensitivity of $log\sigma_0^2$, $log\sigma_1^2$.}]{
  \label{fig:sensitivity}
  \includegraphics[width=0.225\textwidth]{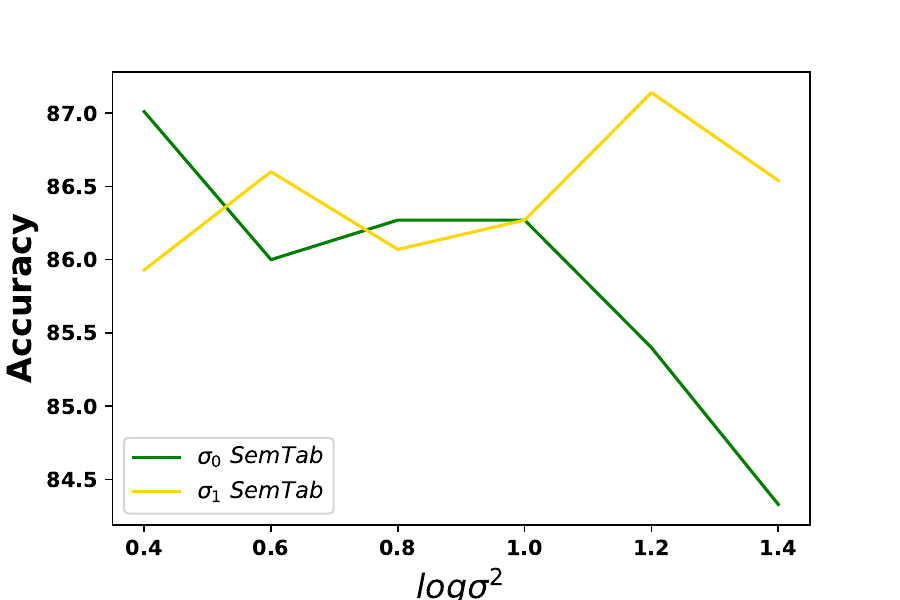}
 }
 \subfigure[\footnotesize{Training curve of $log\sigma_0^2$, $log\sigma_1^2$.}]{
  \label{fig:sigma}
  \includegraphics[width=0.225\textwidth]{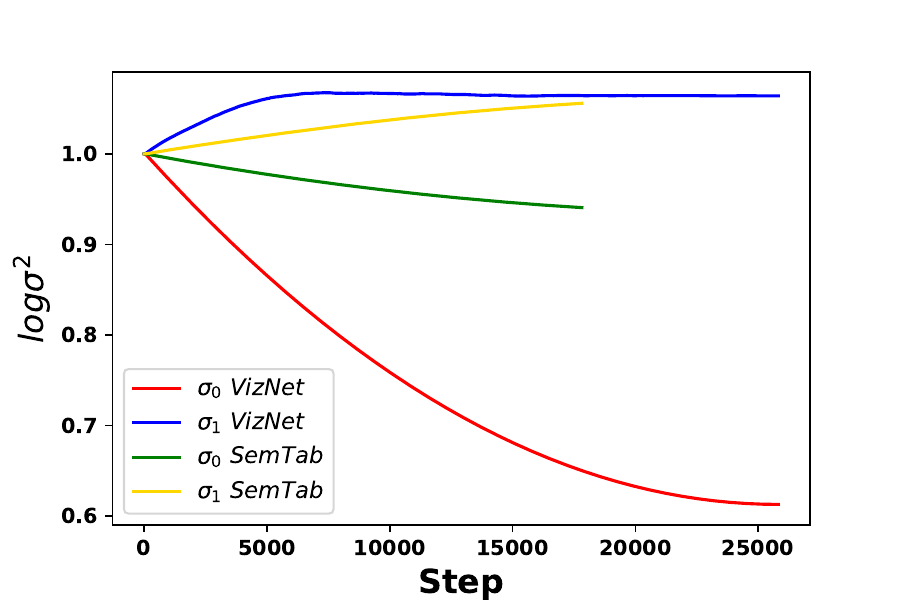}
 }
 \vspace{-0.4cm}
 \caption{Analysis on $\sigma_0$ and $\sigma_1$}
  \vspace{-0.6cm}
\end{figure}
We cut down the original training dataset into a small one to find the suitable data amount to fully benefit from multi-task learning. We introduce $p$ in this experiment, which is the proportion of the training set to the original dataset. For example, when $p=0.2$, the total amount of data would be $0.2$ times the actual amount while the testing set remains unchanged. The experiment result is shown in \autoref{fig:m_w_F1}, we select $p=0.2, 0.4, 0.6, 0.8, 1.0$. When $p$ is small, the training dataset is limited. Hence the model performance would benefit less from the candidate-type representation generation sub-task since the whole model would be more complicated and hard to train.
To illustrate the effectiveness of the row filter, we conducted tests on the KGLink model using filter sizes of $k=10, 25, 50, \text{all}$. These values correspond to using no more than the specified number of rows for each table during annotation. In our experiments, the maximum column encoding length was set to 64, and $all$ indicates retaining at most 64 rows for the input table.

\begin{figure}[t]
 \centering
 \subfigure[\footnotesize{Weighted F1 with varying $k$
 }]{
  \label{fig:w_F1}
  \includegraphics[width=0.22\textwidth]{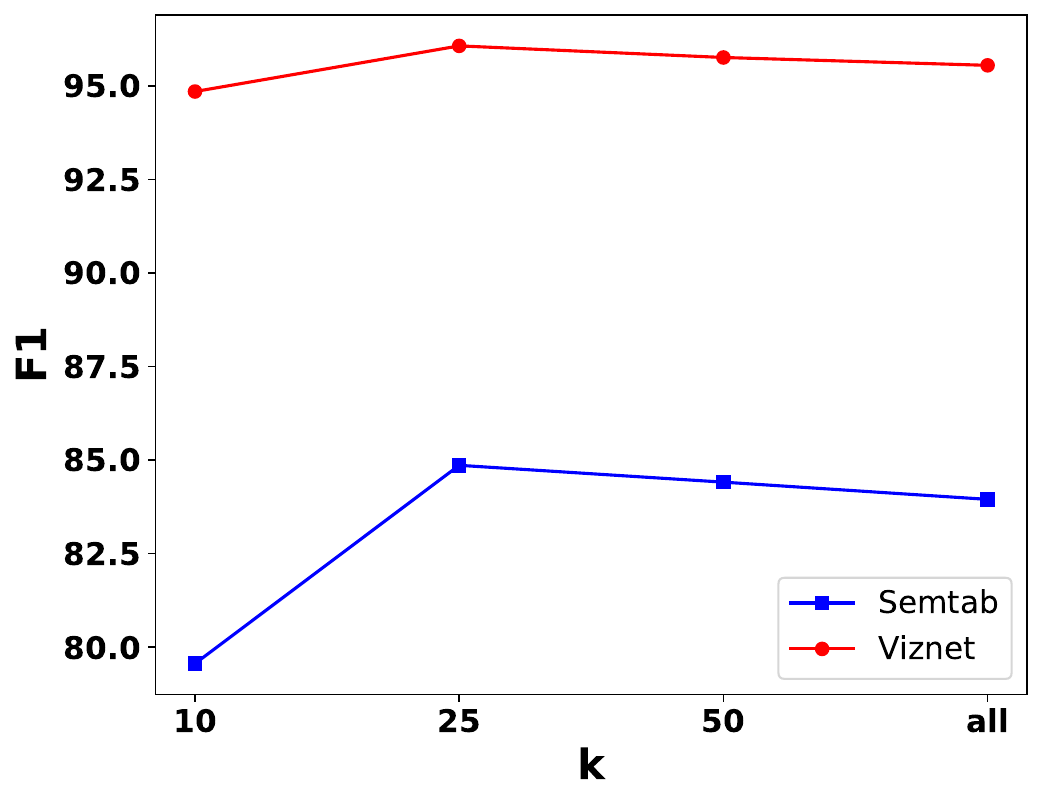}
 }
  \subfigure[\footnotesize{Time cost with varying $k$}]{
  \label{fig:t_F1}
  \includegraphics[width=0.22\textwidth]{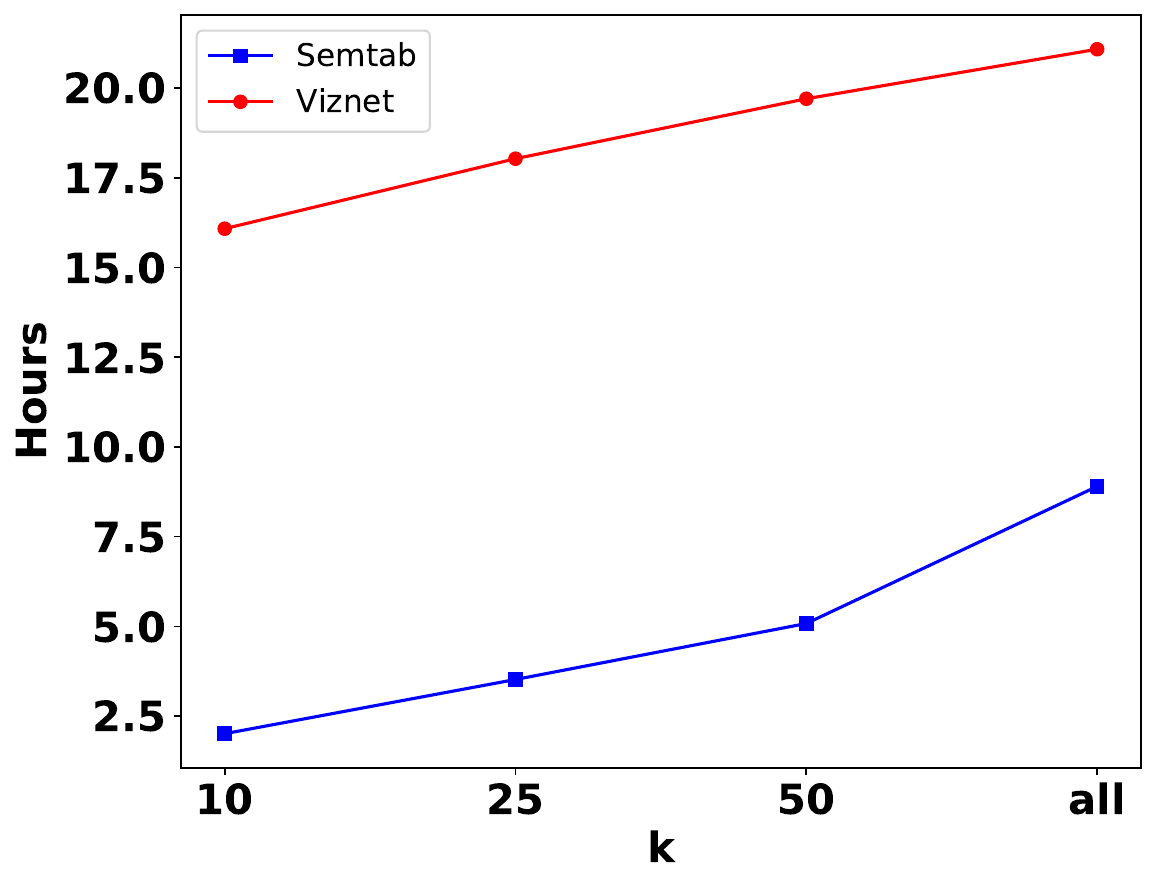}
 }
 \caption{The weighted F1 score and time cost of KGLink with varying $k$}
  \vspace{-0.5cm}
 \label{fig:topkrow}
\end{figure}

As depicted in \autoref{fig:topkrow}, optimal prediction performance is observed when $k$ is set to 25. This suggests that increasing the number of table rows does not necessarily lead to better performance, as additional rows may introduce noise. Therefore, selecting a smaller number of rows that are more suitable for prediction becomes crucial. Additionally, as shown in Figure \autoref{fig:t_F1}, a smaller $k$ maintains better time efficiency, especially on relatively small datasets like SemTab. To strike the right balance between prediction performance and time cost, in our experiments, unless otherwise specified, we set $k=25$.

We present Figure \autoref{fig:sensitivity} here to show the sensitivity of $\sigma_0$ and $\sigma_1$ of the adaptive weighted loss on SemTab dataset, we tested $log\sigma_0^2$ and $log\sigma_1^2$ within range 0.4 to 1.4 (When testing $log\sigma_0^2$, the $log\sigma_1^2$ is fixed to 1), where the accuracy tend to increase with $log\sigma_0^2$ decrease, and 
$log\sigma_0^2$ on the contrary. The range of accuracy in Figure \autoref{fig:sensitivity} demonstrate our model is more sensitive to $\sigma_0$ compared with $\sigma_1$. This shows the need to carefully determine the weight of the column representation generation sub-task, since we do not directly take its result for use, an inappropriate weight may degrade KGLink's performance.
We also present Figure \autoref{fig:sigma} here to show how $log\sigma_0^2$ and $log\sigma_1^2$ are optimized in the two datasets, demonstrating differences in uncertainties in different datasets and tasks.
Experiment on VizNet dataset converged to a smaller $\sigma_0$, which shows that when the dataset contains more tables and fewer column labels, the column representation generation task-related components would be better trained and contribute more to the model.

\vspace{-0.1cm}

\begin{table}[t]
\centering
\caption{Accuracy on test set with no extracted KG information}
\label{tab:no_ct}
\begin{tabular}{l|l|l}
\hline
                                   \textbf{Model}  & \textbf{Numeric Acc}       & \textbf{Non-numeric Acc}        \\ \hline
\multicolumn{1}{l|}{KGLink} &   \textbf{97.04}    & \textbf{90.92} \\ \hline
\multicolumn{1}{l|}{HNN \cite{chen2019learning}} &  44.05   &  18.37 \\ \hline
\multicolumn{1}{l|}{TaBERT \cite{yin2020tabert}} &  96.57   &  90.27 \\ \hline
\multicolumn{1}{l|}{Doduo \cite{suhara2022annotating}} &96.28 & 89.50    \\ \hline
\multicolumn{1}{l|}{RECA \cite{sun2023reca}}   & 96.89 & 61.54 \\ \hline
\multicolumn{1}{l|}{Sudowoodo \cite{wang2022sudowoodo}} & 96.21 & 67.72 \\ \hline
\end{tabular}
\vspace{-0.3cm}
\end{table}

\begin{table}[h]
  \caption{Performance comparison of table filters}
  \label{tab:filter}
  \centering
  \scalebox{0.95}{
\begin{tabular}{c|cc|cc}
\toprule
    \multirow{2}{*}{Filter mechanism}&
	\multicolumn{2}{c}{SemTab}&
	\multicolumn{2}{c}{VizNet} \\
	\cline{2-5}
  & Accuracy & Weighted F1  & Accuracy & Weighted F1 \\
    \midrule
    Our top-$k$ row filter &  \textbf{87.12} & \textbf{85.78}   & \textbf{96.28}  & \textbf{96.07} \\
    Original top-$k$ rows & 85.93 &	84.39  & 96.14 & 95.97  \\ \hline
\end{tabular}}
\vspace{-0.3cm}
\end{table}

\subsection{Data Effciency}

As shown in \autoref{tab:filter}, the original top-$k$ rows mean that we do not apply our row filter and directly select the top-$k$ rows from the table by it's original order. This filter mechanism is deterministic in feature vector generation since it is generated from the cell whose row has the best row linking score, which is used in this filter mechanism.

In the SemTab dataset, the performance improved more significantly compared to the VizNet dataset. This is because the SemTab dataset is more KG-related and has richer KG-extracted information compared to VizNet, as indicated in \autoref{tab:dataset_nolink}. Therefore, it would benefit more from a better filter mechanism.

\subsection{Qualitative Evaluation}\label{QE}
\begin{figure}[t]
 \centering
 \subfigure[\footnotesize{Weighted F1 with varying $p$}]{
  \label{fig:w_F1_p}
  \includegraphics[width=0.22\textwidth]{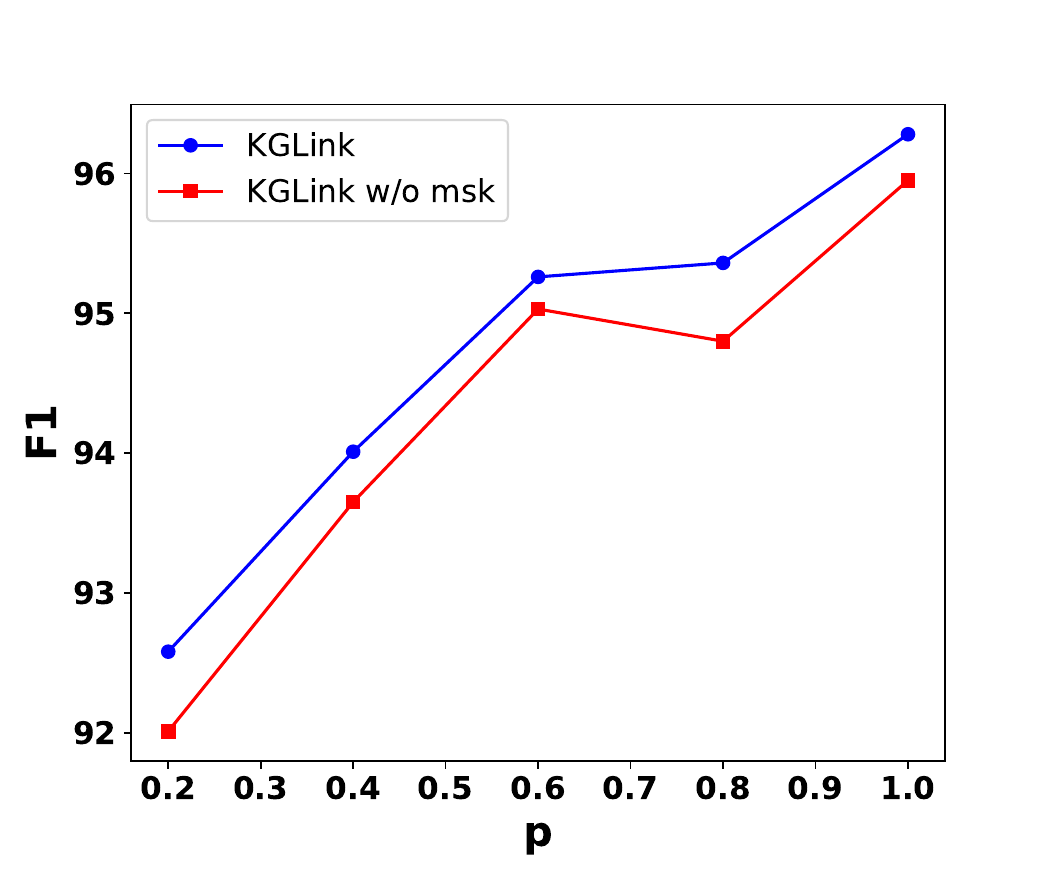}
 }
  \subfigure[\footnotesize{Accuracy with varying $p$}]{
  \label{fig:m_F1}
  \includegraphics[width=0.22\textwidth]{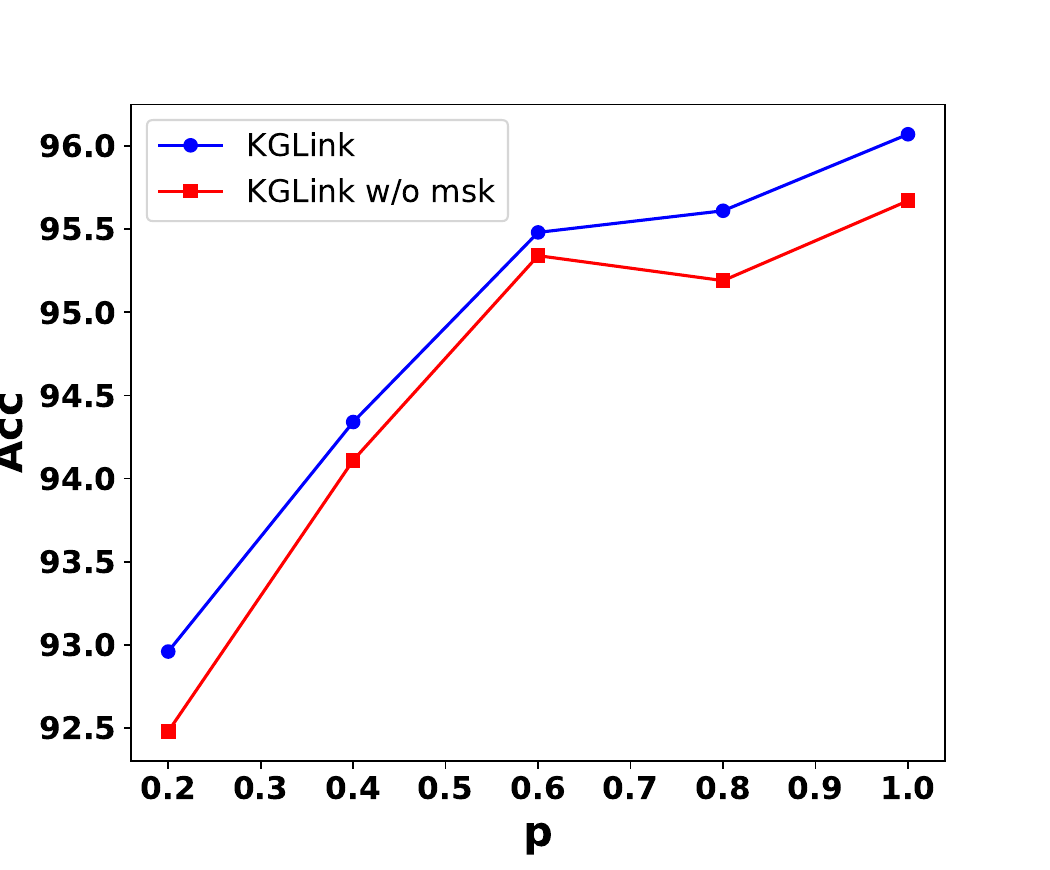}
 }
 \caption{The weighted F1 score and accuracy of KGLink and KGLink $w/o\ msk$ with varying $p$}
  \vspace{-0.5cm}
 \label{fig:m_w_F1}
\end{figure}

For the SemTab dataset, the top three classes with increased accuracy after adding the column representation generation task are: \textit{Athlete}, \textit{Protein}, and \textit{Film}, with an average increase of accuracy of 9.70. It's worth noting that we only consider classes with more than 10 samples in the test set to mitigate randomness.

As depicted in Figure \autoref{fig:tg_issue}, samples from the \textit{Athlete} class often encounter the type granularity issue, indicating that the column representation generation task is beneficial in resolving this problem. For the \textit{Protein} class, we observed that the candidate types extracted from the KG are sometimes not \textit{Protein} itself but could be \textit{Gene} or \textit{Enzymes}, closely related in the semantic space to \textit{Protein}. A similar situation is also observed in the \textit{Film} class, where the candidate types generated from the KG could be \textit{Television series}, \textit{Scholarly article}, or \textit{Book}, instead of \textit{Film}. In such cases, the column representation generation task proves to be beneficial.

For the VizNet dataset, the top three classes with increased accuracy after adding the column representation generation task are: \textit{Artist}, \textit{Year}, and \textit{Rank}, with an average increase of accuracy of 3.18. Notably, since the test set of the VizNet dataset is 10 times larger than that of SemTab, and its label number is only 28\% of SemTab, we only consider classes with more than 100 samples in the test set.

For the \textit{Artist} class, the situation is similar to the \textit{Athlete} class in the SemTab dataset. The information extracted from the KG of this class contains information such as \textit{Human}, \textit{Musical group}, or \textit{Film}, where the column representation generation task is needed to address the granularity issue. The accuracy increase on \textit{Year} and \textit{Code} illustrates that the column representation generation task could also enhance performance on classes containing numeric columns, as we prepend the column's mean, variance, and average value to the column, and these values have a certain range in reality.

\section{Conclusion}
We introduced KGLink, which combines a deep learning PLM with KG information for column-type annotation. Through experiments on benchmark datasets, we demonstrate that KGLink achieves comparable or superior performance compared to state-of-the-art methods. Our analysis highlights the data efficiency of KGLink, achieving similar performance to most of the baseline methods using only approximately 60\% of the training data. To support further research, we have made the KGLink model code and experimental data publicly available.

\section{Acknowledgements}
Lei Chen’s work is partially supported by National Key Research and Development Program of China Grant No. 2023YFF0725100, National Science Foundation of China (NSFC) under Grant No. U22B2060, the Hong Kong RGC GRF Project 16213620, RIF Project R6020-19, AOE Project AoE/E-603/18, Theme-based project TRS T41-603/20R, CRF Project C2004-21G, Guangdong Province Science and Technology Plan Project 2023A0505030011, Hong Kong ITC ITF grants MHX/078/21 and PRP/004/22FX, Zhujiang scholar program 2021JC02X170, Microsoft Research Asia Collaborative Research Grant and HKUST-Webank joint research lab grants.

\clearpage



\begin{thebibliography}{10}
\providecommand{\url}[1]{#1}
\csname url@samestyle\endcsname
\providecommand{\newblock}{\relax}
\providecommand{\bibinfo}[2]{#2}
\providecommand{\BIBentrySTDinterwordspacing}{\spaceskip=0pt\relax}
\providecommand{\BIBentryALTinterwordstretchfactor}{4}
\providecommand{\BIBentryALTinterwordspacing}{\spaceskip=\fontdimen2\font plus
\BIBentryALTinterwordstretchfactor\fontdimen3\font minus \fontdimen4\font\relax}
\providecommand{\BIBforeignlanguage}[2]{{%
\expandafter\ifx\csname l@#1\endcsname\relax
\typeout{** WARNING: IEEEtran.bst: No hyphenation pattern has been}%
\typeout{** loaded for the language `#1'. Using the pattern for}%
\typeout{** the default language instead.}%
\else
\language=\csname l@#1\endcsname
\fi
#2}}
\providecommand{\BIBdecl}{\relax}
\BIBdecl

\bibitem{2021_mtab4wikidata}
\BIBentryALTinterwordspacing
P.~Nguyen, I.~Yamada, N.~Kertkeidkachorn, R.~Ichise, and H.~Takeda, ``Semtab 2021: Tabular data annotation with mtab tool,'' in \emph{SemTab@ISWC 2021}, ser. {CEUR} Workshop Proceedings, vol. 3103.\hskip 1em plus 0.5em minus 0.4em\relax CEUR-WS.org, 2021, pp. 92--101. [Online]. Available: \url{http://ceur-ws.org/Vol-3103/paper8.pdf}
\BIBentrySTDinterwordspacing

\bibitem{sun2023reca}
Y.~Sun, H.~Xin, and L.~Chen, ``Reca: Related tables enhanced column semantic type annotation framework,'' \emph{Proceedings of the VLDB Endowment}, vol.~16, no.~6, pp. 1319--1331, 2023.

\bibitem{wang2022sudowoodo}
R.~Wang, Y.~Li, and J.~Wang, ``Sudowoodo: Contrastive self-supervised learning for multi-purpose data integration and preparation,'' \emph{arXiv preprint arXiv:2207.04122}, 2022.

\bibitem{deng2022turl}
X.~Deng, H.~Sun, A.~Lees, Y.~Wu, and C.~Yu, ``Turl: Table understanding through representation learning,'' \emph{ACM SIGMOD Record}, vol.~51, no.~1, pp. 33--40, 2022.

\bibitem{gormley2015elasticsearch}
C.~Gormley and Z.~Tong, \emph{Elasticsearch: the definitive guide: a distributed real-time search and analytics engine}.\hskip 1em plus 0.5em minus 0.4em\relax " O'Reilly Media, Inc.", 2015.

\bibitem{chen2019learning}
J.~Chen, E.~Jim{\'e}nez-Ruiz, I.~Horrocks, and C.~Sutton, ``Learning semantic annotations for tabular data,'' in \emph{Proceedings of the 28th International Joint Conference on Artificial Intelligence}, 2019, pp. 2088--2094.

\bibitem{nguyenmtab4dbpedia}
P.~Nguyen, N.~Kertkeidkachorn, R.~Ichise, and H.~Takeda, ``Mtab4dbpedia: Semantic annotation for tabular data with dbpedia.''

\bibitem{vaswani2017attention}
A.~Vaswani, N.~Shazeer, N.~Parmar, J.~Uszkoreit, L.~Jones, A.~N. Gomez, {\L}.~Kaiser, and I.~Polosukhin, ``Attention is all you need,'' \emph{Advances in neural information processing systems}, vol.~30, 2017.

\bibitem{robertson2009probabilistic}
S.~Robertson, H.~Zaragoza \emph{et~al.}, ``The probabilistic relevance framework: Bm25 and beyond,'' \emph{Foundations and Trends{\textregistered} in Information Retrieval}, vol.~3, no.~4, pp. 333--389, 2009.

\bibitem{tenney2019bert}
I.~Tenney, D.~Das, and E.~Pavlick, ``Bert rediscovers the classical nlp pipeline,'' \emph{arXiv preprint arXiv:1905.05950}, 2019.

\bibitem{mikolov2013efficient}
T.~Mikolov, K.~Chen, G.~Corrado, and J.~Dean, ``Efficient estimation of word representations in vector space,'' \emph{arXiv preprint arXiv:1301.3781}, 2013.

\bibitem{pennington2014glove}
J.~Pennington, R.~Socher, and C.~D. Manning, ``Glove: Global vectors for word representation,'' in \emph{Proceedings of the 2014 conference on empirical methods in natural language processing (EMNLP)}, 2014, pp. 1532--1543.

\bibitem{liu2019roberta}
Y.~Liu, M.~Ott, N.~Goyal, J.~Du, M.~Joshi, D.~Chen, O.~Levy, M.~Lewis, L.~Zettlemoyer, and V.~Stoyanov, ``Roberta: A robustly optimized bert pretraining approach,'' \emph{arXiv preprint arXiv:1907.11692}, 2019.

\bibitem{clark2019does}
K.~Clark, U.~Khandelwal, O.~Levy, and C.~D. Manning, ``What does bert look at? an analysis of bert's attention,'' \emph{arXiv preprint arXiv:1906.04341}, 2019.

\bibitem{halevy2009unreasonable}
A.~Halevy, P.~Norvig, and F.~Pereira, ``The unreasonable effectiveness of data,'' \emph{IEEE intelligent systems}, vol.~24, no.~2, pp. 8--12, 2009.

\bibitem{hu2019viznet}
K.~Hu, S.~Gaikwad, M.~Hulsebos, M.~A. Bakker, E.~Zgraggen, C.~Hidalgo, T.~Kraska, G.~Li, A.~Satyanarayan, and {\c{C}}.~Demiralp, ``Viznet: Towards a large-scale visualization learning and benchmarking repository,'' in \emph{Proceedings of the 2019 CHI Conference on Human Factors in Computing Systems}, 2019, pp. 1--12.

\bibitem{cafarella2008webtables}
M.~J. Cafarella, A.~Halevy, D.~Z. Wang, E.~Wu, and Y.~Zhang, ``Webtables: exploring the power of tables on the web,'' \emph{Proceedings of the VLDB Endowment}, vol.~1, no.~1, pp. 538--549, 2008.

\bibitem{hulsebos2019sherlock}
M.~Hulsebos, K.~Hu, M.~Bakker, E.~Zgraggen, A.~Satyanarayan, T.~Kraska, {\c{C}}.~Demiralp, and C.~Hidalgo, ``Sherlock: A deep learning approach to semantic data type detection,'' in \emph{Proceedings of the 25th ACM SIGKDD International Conference on Knowledge Discovery \& Data Mining}, 2019, pp. 1500--1508.

\bibitem{yin2020tabert}
P.~Yin, G.~Neubig, W.-t. Yih, and S.~Riedel, ``Tabert: Pretraining for joint understanding of textual and tabular data,'' \emph{arXiv preprint arXiv:2005.08314}, 2020.

\bibitem{chen2019colnet}
J.~Chen, E.~Jim{\'e}nez-Ruiz, I.~Horrocks, and C.~Sutton, ``Colnet: Embedding the semantics of web tables for column type prediction,'' in \emph{Proceedings of the AAAI Conference on Artificial Intelligence}, vol.~33, no.~01, 2019, pp. 29--36.

\bibitem{oliveira2019adog}
D.~Oliveira and M.~d'Aquin, ``Adog-annotating data with ontologies and graphs,'' in \emph{SemTab@ ISWC}, 2019.

\bibitem{abdelmageed2020jentab}
N.~Abdelmageed and S.~Schindler, ``Jentab: Matching tabular data to knowledge graphs.'' in \emph{SemTab@ ISWC}, 2020, pp. 40--49.

\bibitem{GDS}
``Google data studio,'' 2020.

\bibitem{MPBI}
``Microsoft power bi,'' 2020.

\bibitem{tableau}
``Tableau software,'' 2020.

\bibitem{iida2021tabbie}
H.~Iida, D.~Thai, V.~Manjunatha, and M.~Iyyer, ``Tabbie: Pretrained representations of tabular data,'' \emph{arXiv preprint arXiv:2105.02584}, 2021.

\bibitem{suhara2022annotating}
Y.~Suhara, J.~Li, Y.~Li, D.~Zhang, {\c{C}}.~Demiralp, C.~Chen, and W.-C. Tan, ``Annotating columns with pre-trained language models,'' in \emph{Proceedings of the 2022 International Conference on Management of Data}, 2022, pp. 1493--1503.

\bibitem{lecun2015deep}
Y.~LeCun, Y.~Bengio, and G.~Hinton, ``Deep learning,'' \emph{nature}, vol. 521, no. 7553, pp. 436--444, 2015.

\bibitem{devlin2018bert}
J.~Devlin, M.-W. Chang, K.~Lee, and K.~Toutanova, ``Bert: Pre-training of deep bidirectional transformers for language understanding,'' \emph{arXiv preprint arXiv:1810.04805}, 2018.

\bibitem{hinton2015distilling}
G.~Hinton, O.~Vinyals, J.~Dean \emph{et~al.}, ``Distilling the knowledge in a neural network,'' \emph{arXiv preprint arXiv:1503.02531}, vol.~2, no.~7, 2015.

\bibitem{kendall2018multi}
A.~Kendall, Y.~Gal, and R.~Cipolla, ``Multi-task learning using uncertainty to weigh losses for scene geometry and semantics,'' in \emph{Proceedings of the IEEE conference on computer vision and pattern recognition}, 2018, pp. 7482--7491.

\bibitem{wang2020learning}
Q.~Wang, L.~Yang, B.~Kanagal, S.~Sanghai, D.~Sivakumar, B.~Shu, Z.~Yu, and J.~Elsas, ``Learning to extract attribute value from product via question answering: A multi-task approach,'' in \emph{Proceedings of the 26th ACM SIGKDD International Conference on Knowledge Discovery \& Data Mining}, 2020, pp. 47--55.

\bibitem{wang2021tcn}
D.~Wang, P.~Shiralkar, C.~Lockard, B.~Huang, X.~L. Dong, and M.~Jiang, ``Tcn: Table convolutional network for web table interpretation,'' in \emph{Proceedings of the Web Conference 2021}, 2021, pp. 4020--4032.

\bibitem{hassanzadeh_oktie_2019_3518539}
\BIBentryALTinterwordspacing
O.~Hassanzadeh, V.~Efthymiou, J.~Chen, E.~Jiménez-Ruiz, and K.~Srinivas, ``{SemTab 2019: Semantic Web Challenge on Tabular Data to Knowledge Graph Matching Data Sets},'' Oct. 2019. [Online]. Available: \url{https://doi.org/10.5281/zenodo.3518539}
\BIBentrySTDinterwordspacing

\bibitem{zhang2020sato}
\BIBentryALTinterwordspacing
D.~Zhang, Y.~Suhara, J.~Li, M.~Hulsebos, {\c{C}}.~Demiralp, and W.-C. Tan, ``Sato: Contextual semantic type detection in tables,'' \emph{Proc. VLDB Endow.}, vol.~13, no.~12, p. 1835–1848, 2020. [Online]. Available: \url{https://doi.org/10.14778/3407790.3407793}
\BIBentrySTDinterwordspacing

\bibitem{khurana2021semantic}
U.~Khurana and S.~Galhotra, ``Semantic concept annotation for tabular data,'' in \emph{Proceedings of the 30th ACM International Conference on Information \& Knowledge Management}, 2021, pp. 844--853.

\bibitem{vrandevcic2014wikidata}
D.~Vrande{\v{c}}i{\'c} and M.~Kr{\"o}tzsch, ``Wikidata: a free collaborative knowledgebase,'' \emph{Communications of the ACM}, vol.~57, no.~10, pp. 78--85, 2014.

\bibitem{lehmann2015dbpedia}
J.~Lehmann, R.~Isele, M.~Jakob, A.~Jentzsch, D.~Kontokostas, P.~N. Mendes, S.~Hellmann, M.~Morsey, P.~Van~Kleef, S.~Auer \emph{et~al.}, ``Dbpedia--a large-scale, multilingual knowledge base extracted from wikipedia,'' \emph{Semantic web}, vol.~6, no.~2, pp. 167--195, 2015.

\bibitem{paszke2019pytorch}
A.~Paszke, S.~Gross, F.~Massa, A.~Lerer, J.~Bradbury, G.~Chanan, T.~Killeen, Z.~Lin, N.~Gimelshein, L.~Antiga \emph{et~al.}, ``Pytorch: An imperative style, high-performance deep learning library,'' \emph{Advances in neural information processing systems}, vol.~32, 2019.

\bibitem{wolf2020transformers}
T.~Wolf, L.~Debut, V.~Sanh, J.~Chaumond, C.~Delangue, A.~Moi, P.~Cistac, T.~Rault, R.~Louf, M.~Funtowicz \emph{et~al.}, ``Transformers: State-of-the-art natural language processing,'' in \emph{Proceedings of the 2020 conference on empirical methods in natural language processing: system demonstrations}, 2020, pp. 38--45.

\bibitem{spacy2}
M.~Honnibal and I.~Montani, ``{spaCy 2}: Natural language understanding with {B}loom embeddings, convolutional neural networks and incremental parsing,'' 2017, to appear.

\bibitem{wang2022language}
T.~Wang, A.~Roberts, D.~Hesslow, T.~Le~Scao, H.~W. Chung, I.~Beltagy, J.~Launay, and C.~Raffel, ``What language model architecture and pretraining objective works best for zero-shot generalization?'' in \emph{International Conference on Machine Learning}.\hskip 1em plus 0.5em minus 0.4em\relax PMLR, 2022, pp. 22\,964--22\,984.

\bibitem{raffel2020exploring}
C.~Raffel, N.~Shazeer, A.~Roberts, K.~Lee, S.~Narang, M.~Matena, Y.~Zhou, W.~Li, and P.~J. Liu, ``Exploring the limits of transfer learning with a unified text-to-text transformer,'' \emph{The Journal of Machine Learning Research}, vol.~21, no.~1, pp. 5485--5551, 2020.

\bibitem{he2020deberta}
P.~He, X.~Liu, J.~Gao, and W.~Chen, ``Deberta: Decoding-enhanced bert with disentangled attention,'' \emph{arXiv preprint arXiv:2006.03654}, 2020.

\end{thebibliography}
\end{document}